\documentclass[manuscript]{acmart}

\usepackage{todonotes}

\usepackage{amsmath}

\DeclareMathOperator*{\argmin}{arg\,min}

\AtBeginDocument{%
  \providecommand\BibTeX{{%
    \normalfont B\kern-0.5em{\scshape i\kern-0.25em b}\kern-0.8em\TeX}}}

\acmJournal{TOSN}

\begin{document}

\title{Underwater Acoustic Networks for Security Risk Assessment in Public Drinking Water Reservoirs}

\author{Jörg Stork}
\affiliation{%
  \institution{TH Köln}
  \streetaddress{Steinmüllerallee 1}
  \city{Gummersbach}
  \state{NRW}
  \country{Germany}
  \postcode{51643}
}

\author{Philip Wenzel}
\affiliation{%
  \institution{TH Köln}
  \streetaddress{Steinmüllerallee 1}
  \city{Gummersbach}
  \state{NRW}
  \country{Germany}
  \postcode{51643}
}

\author{Severin Landwein}
\affiliation{%
  \institution{TH Köln}
  \streetaddress{Steinmüllerallee 1}
  \city{Gummersbach}
  \state{NRW}
  \country{Germany}
  \postcode{51643}
}

\author{Maria-Elena Algorri}
\email{elena.algorri@th-koeln.de}
\affiliation{%
  \institution{TH Köln}
  \streetaddress{Steinmüllerallee 1}
  \city{Gummersbach}
  \state{NRW}
  \country{Germany}
  \postcode{51643}
}

\author{Martin Zaefferer}
\affiliation{%
  \institution{Bartz \& Bartz GmbH}
  \streetaddress{Goebenstr. 10}
  \city{Gummersbach}
  \state{NRW}
  \country{Germany}
  \postcode{51643}
}

\author{Wolfgang Kusch}
\affiliation{%
  \institution{Aggerverband}
  \streetaddress{Sonnenallee 40}
  \city{Gummersbach}
  \state{NRW}
  \country{Germany}
  \postcode{51645}
}

\author{Martin Staubach}
\affiliation{%
  \institution{Aggerverband}
  \streetaddress{Sonnenallee 40}
  \city{Gummersbach}
  \state{NRW}
  \country{Germany}
  \postcode{51645}
}

\author{Thomas Bartz-Beielstein}
\affiliation{%
  \institution{TH Köln}
  \streetaddress{Steinmüllerallee 1}
  \city{Gummersbach}
  \state{NRW}
  \country{Germany}
  \postcode{51643}
}

\author{Hartmut Köhn}
\affiliation{%
  \institution{TH Köln}
  \streetaddress{Steinmüllerallee 1}
  \city{Gummersbach}
  \state{NRW}
  \country{Germany}
  \postcode{51643}
}

\author{Hermann Dejager}
\affiliation{%
  \institution{TH Köln}
  \streetaddress{Steinmüllerallee 1}
  \city{Gummersbach}
  \state{NRW}
  \country{Germany}
  \postcode{51643}
}

\author{Christian Wolf}
\affiliation{%
  \institution{TH Köln}
  \streetaddress{Steinmüllerallee 1}
  \city{Gummersbach}
  \state{NRW}
  \country{Germany}
  \postcode{51643}
}

\renewcommand{\shortauthors}{Stork et al.}

\begin{abstract}
We have built a novel system for the surveillance of drinking water reservoirs using underwater sensor networks. We implement an innovative AI-based approach to detect, classify and localize underwater events. In this paper, we describe the technology and cognitive AI architecture of the system based on one of the sensor networks, the hydrophone network. We discuss the challenges of installing and using the hydrophone network in a water reservoir where traffic, visitors, and variable water conditions create a complex, varying environment. Our AI solution uses an autoencoder for unsupervised learning of latent encodings for classification and anomaly detection, and time delay estimates for sound localization. Finally, we present the results of experiments carried out in a laboratory pool and the water reservoir and discuss the system's potential.
\end{abstract}



\maketitle

\section{Introduction}\label{sec:intro}
Public drinking water reservoirs are an essential part of public infrastructure. Next to drinking water supply and renewable energy generation, they are also used to regulate the water flow in rivers and are frequently used for recreational activities. According to the Federal Office for Civil Protection and Disaster Assistance and the Federal Office for Information Security in Germany, public drinking water reservoirs are an essential part of the critical infrastructure due to their size and a considerable amount of stored water. In the most recent evaluation of the security of the drinking water supply in Germany~\cite{BBK19b} drinking water reservoirs were identified as one possible target for terrorist attacks that have a huge damage potential. This confirms the findings of the reports on Terrorism and Security Issues Facing the Water Infrastructure Sector~\cite{CRS10b} and the Dam Safety Report by the Congressional Research Service of the United States of America~\cite{CRS19b} stating that 17\% of all dams in the USA have a high hazard potential with a high probability for loss of life. Therefore, comprehensive monitoring of dam structures is crucial in identifying potential security threats as soon as possible to maximize the time for preventive measures such as the evacuation of populated areas. As many different kinds of drinking water reservoir dams exist (concrete dams, prestressed concrete dams, counterfort type, and combination type dams), the respective monitoring solutions need to be adapted to the dam type and also to the dam type-specific threats. 

Currently available and commonly used solutions for dam monitoring focus on structural damage detection in and outside of the dam ~\cite{multiconsult21b} but until now not on monitoring the movement activity underwater in the closer dam vicinity that might pose a potential threat. Furthermore, analysis of measured monitoring data is often not automated and relies on conventional statistical methods. According to ~\cite{li19b} analysis of dam monitoring data can be clustered into the following three categories: (1) monitoring models, (2) monitoring index, and (3) abnormal value detection. Particularly for model building and abnormal value detections, the use of machine learning methods has become widely popular due to the high amount of historical data available for dams. The purpose of this paper is to present a comprehensive sensor network solution for the monitoring of dams of water reservoirs at the waterside and beneath the water surface, which is part of the research project “Safety of water reservoir dams” funded by the German Ministry for Education and Research. The network consists of three different components: acoustical sensors - hydrophones, sonars for localization and tracking, as well as an underwater robot drone for on-site inspection. The data is gathered in the dam on-site and analyzed using AI techniques, such as feature identification and models for classification and anomaly detection. 

Sound recognition is frequently, but not exclusively, applied to speech and music. In ~\cite{Chachada2014} the authors give an overview of the application of sound recognition to environmental sounds, which partially matches to the types of sounds that occur in the context of our application. Bayram et al.~\cite{bayram2021real} utilize a sequential Autoencoder to detect acoustic anomalies in an industrial process with focus on a real-time implementation. Marchi et al.~\cite{marchi2015novel} describe an unsupervised approach based on a denoising bidirectional Long Short-Term Memory (LSTM) recurrent Autoencoder.

More closely related to the field of security applications, Radhakrishnan et al.~\cite{Radhakrishnan2005} and Atrey et al.~\cite{Atrey2006} use sounds signals to perform event detection for surveillance. Shipps and Abraham~\cite{Shipps2004} describe how acoustic signals can be employed to monitor underwater security, such as required in waterways and ports.

The remainder of this paper is organized as follows. Section 2 describes the sensors and their installation at the dam is detailed in Section 3, while Section 4 focuses on data acquisition, transmission, and storage. The AI software infrastructure to detect security threats is presented in Section 5, and their application and initial results are given in the form of real-life case studies in Section 6. Sections 7 and 8 discuss and sum up the major contributions and give an outlook on future research.

\section{Sensors and Physical Considerations }\label{sec:sensors}

\subsection{Hydrophones}
The acoustic sensor used to record underwater sounds was the spherical hydrophone ASF-1 MKII from Ambient. 
The frequencies, which the hydrophone can record, range from 7Hz to 40kHz. 
The hydrophone provides omnidirectional response over its whole bandwidth, 
thus ensuring that the direction from which noise comes does not affect its recorded volume. 

We deployed an acoustic network consisting of five hydrophones at well-defined locations inside the water dam. 
The hydrophone network allows us to acquire underwater sounds and also to perform sound localization. 
To achieve accurate sound localization, 
hardware considerations must be taken to avoid problems of signal synchronization when measuring time delays in sound recordings. This led us to connect all five hydrophones to a single multi-channel audio amplifier using XLR connectors instead of using individual audio amplifiers for each hydrophone. 
Using a single multi-channel audio amplifier, 
we can simultaneously record the data from all the hydrophones. 
The disadvantage of our hardware design was that we needed much longer acoustic cables and faced additional challenges with the system's physical installation. 

\subsection{Multi-channel Audio amplifier }
We use the U-Phoria UMC 1820 multi-channel audio amplifier from Behringer. 
The amplifier provides the hydrophones with the required 48V phantom power supply through cables that can be up to 200m long. The amplifier's main task is to convert the analog audio signals of the hydrophones into digital ones. 
It can scan up to 8 XLR audio inputs with a sampling rate of 96 kHz at a resolution of 24-bit. 
The five hydrophones in our network are scanned in parallel, and each one is mapped to a separate audio channel. 
For real-time analysis of the hydrophone data, we connect the audio amplifier via a USB interface to a computer running Ubuntu 18.04. 

\section{Installation}\label{sec:install}

The sensor system consists of the network of hydrophones installed underwater and the signal acquisition computers located in the switchboard area inside the barrage (the wall of the dam). The sensor system should be highly available in the rough ambient conditions of the dam. Additionally, the underwater hydrophones should be serviced efficiently and at a low cost; that is, it should be possible to bring them to the water surface without divers. For these purposes, we tested two installation options for the hydrophones: Direct installation at the barrage (dam wall) and installation in the open water in front of the barrage.
 
\begin{figure}[h]
  \centering
  \includegraphics[width=0.99\linewidth]{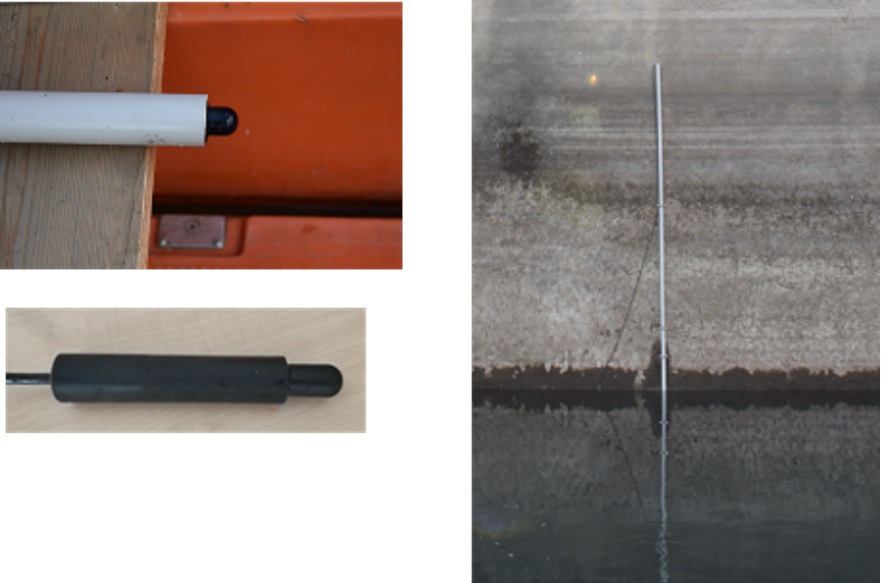}
  \caption{
  Upper Left: Hydrophone inserted in a plastic pipe. 
  Lower Left: The hydrophone.
  Right: The installation of the hydrophone on the barrage.}
  \label{fig:hydro}
\end{figure}

The installation of the hydrophones directly on the barrage has the advantage that there are no cables in the open water that could interfere with other sensors. In addition, the cable lengths from the hydrophones to the audio amplifier located inside the barrage can be kept as short as possible. A disadvantage of this installation type is the acoustic coupling of the hydrophones with the barrage. This is a problem because of the public road that runs along with the barrage. Every vehicle driving along the barrage causes an individual noise that depends on its weight, size, tires, and speed. The traffic noise is superimposed as undesired acoustic data. In general, placing the hydrophones next to a hard surface (barrage or dam bottom) reduces their sensitivity to low-level underwater noises and compromises the dam structure's monitoring.
Figure~\ref{fig:hydro} shows the installation of the hydrophones on the barrage. 
In the initial construction, the hydrophones were inserted in a plastic tube with a bearing (Fig.~\ref{fig:hydro} upper left), which we then changed into a stainless-steel pipe suspended with rubber vibration dampers to minimize the acoustic coupling to the barrage. 

The use of vibration dampers on the hydrophones to reduce the mechanical coupling and thus the adjacent road's acoustic artifacts was unsuccessful. The barrage is such a large resonator that the water layers in front of it transmit the barrage's vibrations, and these vibrations are then indirectly (and directly) recorded by the hydrophones. For this reason, we decided to install the hydrophones at a distance from the barrage. Extensive experiments with different positions showed that the hydrophones must be installed at least 10m away from the barrage to reduce traffic noise to a tolerable level. 

This installation of the hydrophones must fulfill the following requirements: They must be placed deep enough to isolate them from disruptive external factors like vandalism, wave beat, currents, and weather; their immersion should be variable to allow for maintenance and because the tidal range of the dam (variation in water depth over the year) is about 13m; finally variations in the depth of the hydrophones allow the reconfiguration of the sensor network. 

\begin{figure}[h]
  \centering
  \includegraphics[width=0.5\linewidth]{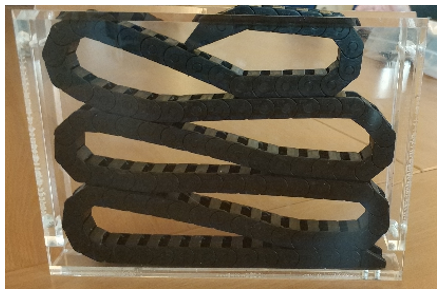}
  \caption{Energy chain system for routing the audio cables from the dam into the barrage without tangling. The energy chain is packed inside a metal gitter at the bottom of the buoy-pulley construction. It is shown here inside an acrylic box for illustration purposes.}
  \label{fig:I1}
\end{figure}

\begin{figure}[h]
  \centering
  \includegraphics[width=0.5\linewidth]{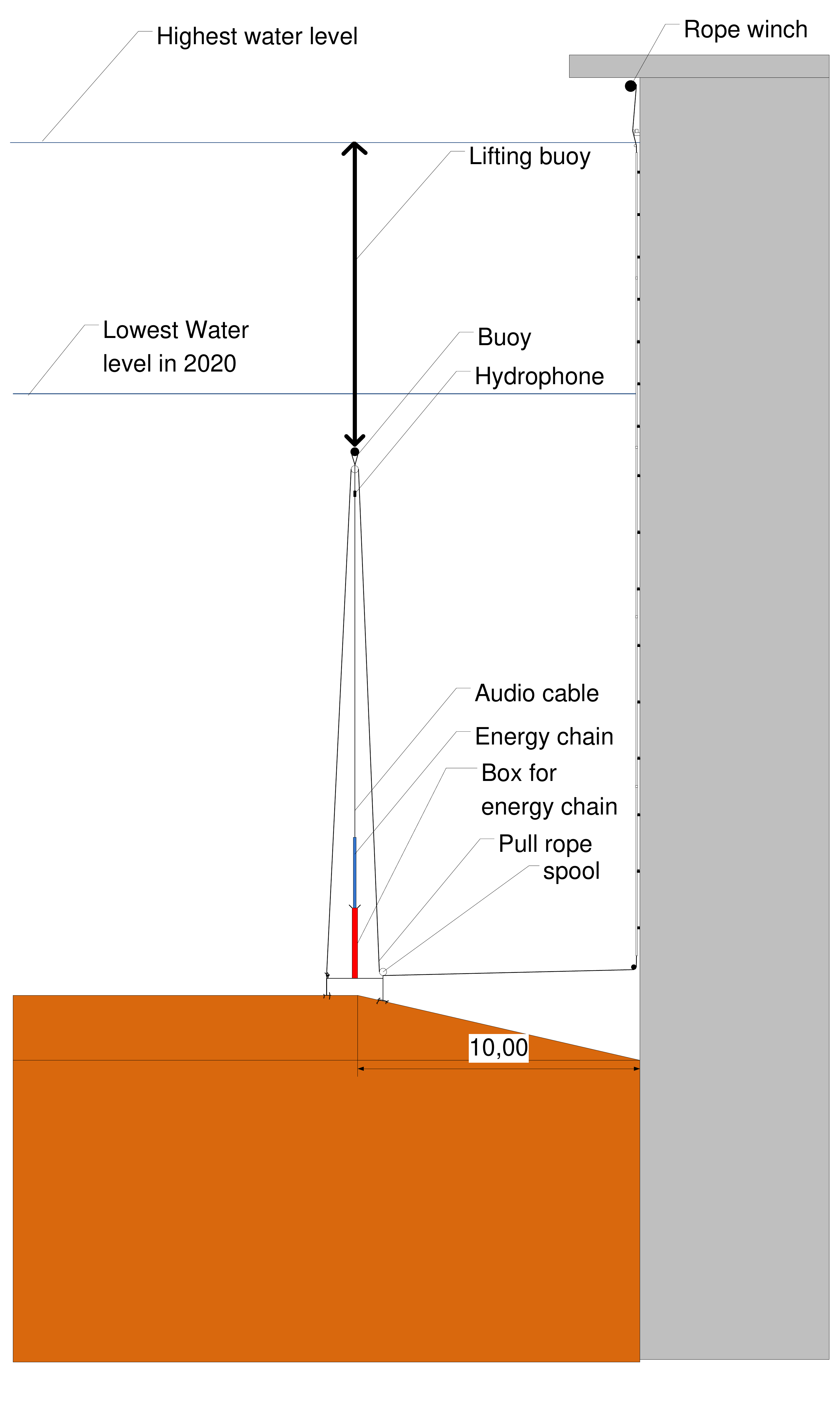}
  \caption{Buoy-hydrophone-construction for installation in dams.}
  \label{fig:I2}
\end{figure}

A buoy-rope-construction with a fixed pulley on the ground and a movable pulley at the buoy was used for the installation. In this double-way, half-strength construction, the hydrophones are fixed to the buoy that can be pulled underwater for normal working conditions and brought to the surface for maintenance.
A critical aspect of the installation was handling the pull cables of the buoy-rope construction and the audio cables of the hydrophones. Both cables can be more than 100m long and tangle easily with the underwater currents causing acoustic noise, mechanical stress, and endangering the underwater robots that move inside the dam. To avoid cable tangle and ensure the orderly routing of the cables to the barrage, the audio cables are guided inside an energy chain. The energy chain is placed inside a metal bar grid at the bottom of the construction. Figure~\ref{fig:I1} shows the energy chain inside an acrylic box. Figure~\ref{fig:I2} shows the location of the energy chain in the buoy-rope construction.

\section{Data Acquisition, Transmission and Storage}\label{sec:data}
\subsection{Data Acquisition}
The acoustic data is input into the computer through the multi-channel audio amplifier that samples the analog audio of the hydrophones at 96KHz and 24 bits per sample. 
We use the Python PyAudio library to record, analyze, and store the digital audio data.  
PyAudio provides Python bindings for PortAudio, a cross-platform audio I/O library.  
PortAudio can open the individual audio channels of the five hydrophones as data streams.  

The data acquisition program is event-based, and PyAudio works in callback mode. 
Data is continuously recorded and cached by PyAudio in a buffer of size $S$. 
As soon as the buffer is full, a callback function that defines how the data is processed is triggered, and PyAudio empties the buffer.
We empirically set the size $S$ of the buffer to six seconds because: 
1) We wanted a buffer size that is long enough to allow us to analyze persistent acoustic events in a block and 
2) we needed a buffer size that is short enough to allow for near real-time data analysis and alarms.  
Consequently, near real-time event alarms have a delay of approximately 6s, and data processing algorithms work on 6s data segments. 
The byte size of the data segments $S$ is given by: $S$ = sampling rate $\times$ \text{bitsPerSample} $\times$ duration = 96K $\times$ 24 $\times$ 6 = 13.8 Mbytes. 

\subsection{Data Transmission}\label{sec:transmission}
\begin{figure}[h]
 \centering
 \includegraphics[width=0.99\linewidth]{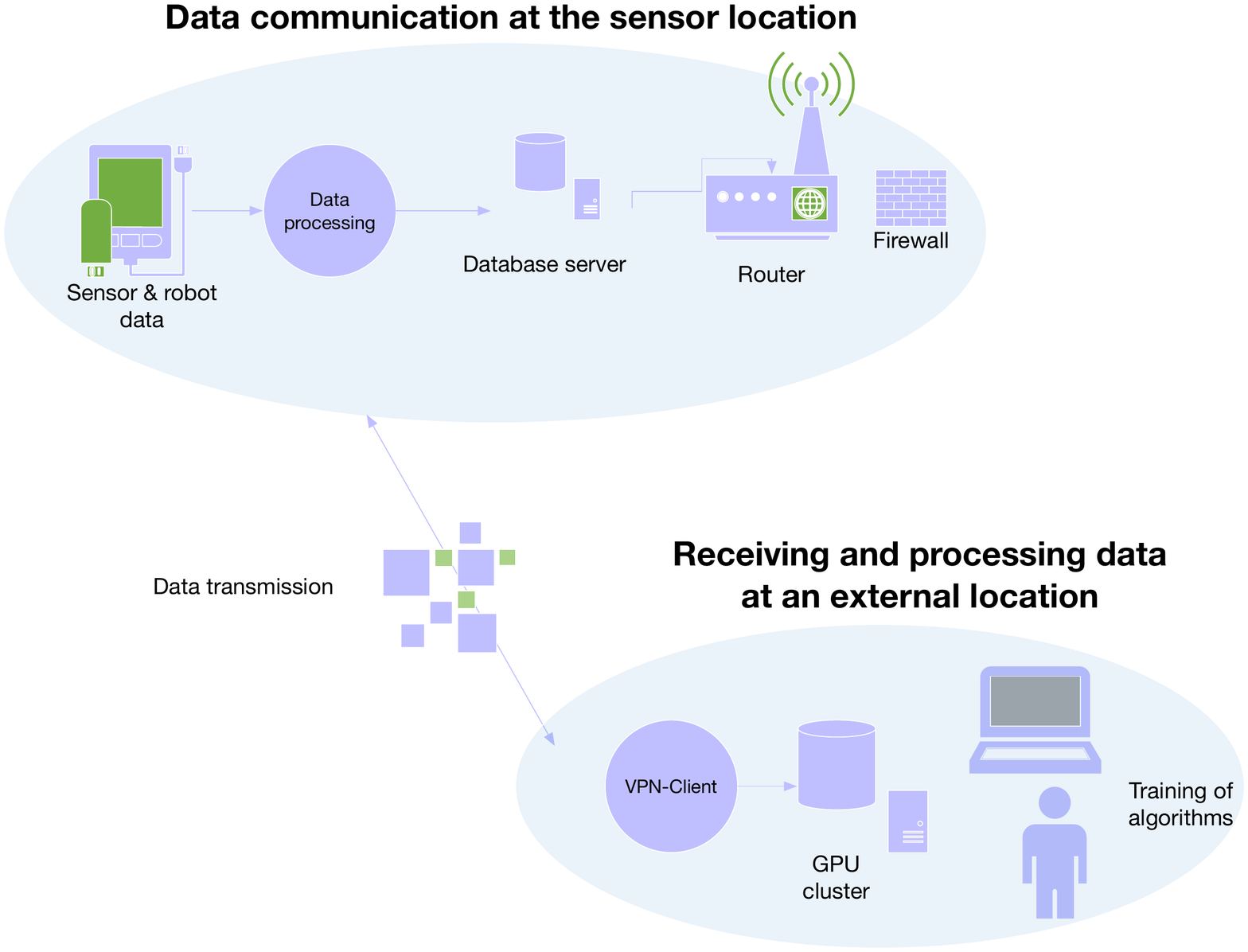}
 \caption{Schematic structure of data transmission. Our hardware structures are divided into two sites, one directly connected to the sensors and responsible for immediate processing (using trained data models) and storing all incoming data. The other is an external HPC cluster to train the respective data models based on the stored data. The sites exchange data and the trained models.}
 \label{fig:coms}
\end{figure}

Figure~\ref{fig:coms} shows the structural design of the data transmission from the sensor to the data evaluation. 
The hydrophones are permanently installed at the dam and transmit their data via cable connection into the dam. 
An industrial PC was installed there, which processes the data in the first step and then forwards it via fiber optic cable / Ethernet to the NAS system (database server), 
where the data is stored and mirrored to maintain data consistency after possible drive failures. 
Access to the data from outside is controlled and regulated by a hardware firewall. 
Access to the network is only possible through a VPN client. 
The data can then be accessed through a VPN tunnel and used for further processing. 
Data security can thus be guaranteed to a high degree. 

For data storage, the relational database management system MariaDB was selected and installed on the NAS server. 
The data of the individual measuring hydrophones is stored there.  

The data is transferred to a remote site to train the AI algorithms,
where a suitable GPU cluster with enough computing power is available to train our models. 
After the training is completed, the learned data models are transferred to the database system installed on-site. 
In the second stage of this project, the now outsourced data evaluation will take place directly on site, as the already learned models will require less computer power to reinforce their learning continuously.

\section{AI Solution for detection of underwater security threats}
The detection of underwater security threats is based on a modular artificial intelligence pipeline. The AI solution's primary goal is the robust identification of any security risks based on the data gathered by the underwater sensor network, in this case, on the audio streams of the distributed hydrophones. Following the identification, the system must immediately deliver feedback to the users and notify them in case of a discovered security risk. 

Three main challenges drive the development of the complete system: 

\begin{enumerate}
\item Recognition and security assessment of unknown event classes. 
The system needs to be able to detect new, previously unobserved event classes and ask for immediate human interaction and feedback to assess and reduce the security risk level. 
\item The recognition of events with a known risk-level, i.e., 
the correct classification of incoming data which belongs to an already observed event-class and immediate notification of the user,
if an event of a high-security risk class emerges.  
\item Localization of sound events. 
The localization of sound sources enables to estimate of the security risk level of a classified event. 
If dangerous activities are identified near the dam, the system needs to notify the user immediately.  
\end{enumerate}

These challenges require a highly dynamic approach, 
which can frequently update and improve the AI models with the acquired data. 
As the initially available amount of labeled and known data is sparse, the system requires a combination of supervised and unsupervised learning techniques. 

Hence, we developed algorithms for the dedicated tasks of 
classification, anomaly detection, and sound localization, 
to tackle all the above-stated challenges. 
The algorithms are connected in a data recognition pipeline and used as an ensemble to allow the desired robust classification by aggregating their results. 
The complete pipeline is further embedded into a software platform, 
which manages the pipeline and establishes user feedback and interaction.  
Examples for the application are given in Section~\ref{sec:experiments}, where a use case for each algorithm is presented.  

\subsection{Data Preprocessing}\label{ssec:prepro}
\begin{figure}[h]
 \centering
 \includegraphics[scale = 0.3]{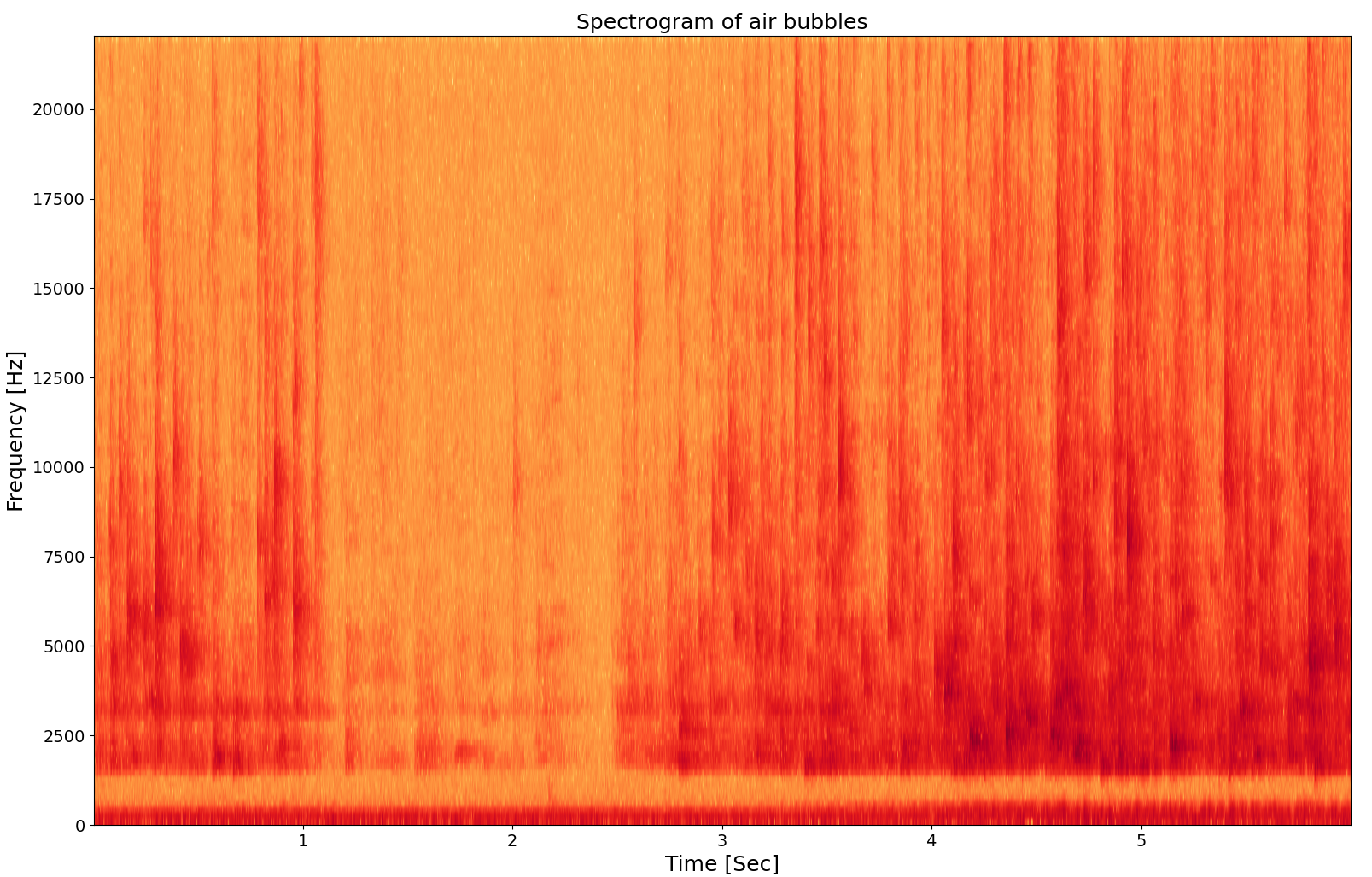}
 \caption{
 Spectrogram of the temporal changes in the frequency content of the audio data produced by air bubbles inside the dam.}
 \label{fig:spec}
\end{figure}

We analyze the audio data in the frequency domain to characterize and identify acoustic events' source position. 
Due to the rapid temporal changes that can occur in the frequency spectrum if a sound is concise, 
we use the short-time Fourier transformation (STFT) for the analysis. 
We use time windows of 100ms to perform the STFT along the audio streams. 
Considering a sampling rate of 96KHz, this time-window length allows us to resolve frequencies between 10Hz 
(the lowest frequency that passes in a 100ms window, 1 / 100ms) 
and 48kHz 
(the largest frequency that can be sampled at 96kHz, 96kHz / 2.0). 
This also matches the capability of the hydrophones, whose frequency range goes from 7Hz to 40kHz. 
To ensure a continuous analysis of the acoustic data's spectral footprint, 
we allow for a 50\% time overlap between consecutive time windows. 
This means that time-window 1 starts at 0ms and ends at 100ms 
while time window 2 starts at 50ms and ends at 150ms. 
Spectrograms are used to visualize the temporal changes in the frequency spectrum of the audio streams. 
Figure~\ref{fig:spec} shows the spectrogram of the noise produced by air bubbles going from the bottom to the top of the dam. 
The spectrogram shows audio-energy in the range of 1KHz to approx. 20KHz. 

We use the FFT-spectrograms for sound localization.
For the classification of sound sources and anomaly detection, we convert the spectrograms to the Mel-spectrum as follows: First, the power spectra are converted to dB-scaling with the max being 0 dB noise level. 
The data is then clipped above –10dB and below –80dB to soften the influence of silent signal noise and loud artifacts. 
Afterwards, a fixed number of 128 Mel frequency bands are computed, and each band is further normalized to [-1:1] intervals,
which is beneficial for the processing with artificial neural networks (ANNs). 
Mel-spectrograms are commonly used for acoustic data, including acoustic data analysis with Autoencoders~\cite{Amir17a}.

\subsection{Autoencoder for Generating Sequence-Encodings }
Based on these Mel-spectra, we utilize an Autoencoder (AE) to generate a latent representation, 
i.e., data encodings for each audio sequence, 
which are utilized as features for the classification algorithm. 
We utilize an AE for the task of unsupervised learning of data encodings 
$f \in \mathbb{R}^p = \mathbf{F}$
which represent the underlying data 
$x \in \mathbb{R}^d = \mathbf{X}$. 
The AE is trained utilizing a structure where an encoder 
$\phi: \mathbf{X} \rightarrow \mathbf{F}$
competes with a decoder 
$\psi: \mathbf{F} \rightarrow \mathbf{X}$. 
The target is to reduce the loss: 
$\phi, \psi = \argmin ||\mathbf{X} - (\psi \circ \phi) \mathbf{X}||$.

Based on these Mel-spectra, we train an Autoencoder (AE) model 
to generate a latent representation based on unsupervised learning, 
i.e., data encodings for each audio sequence. 
The data encodings represent the underlying data and serve as input features for the classification algorithm. 
The AE is a structured neural network with several layers where an encoder competes with a decoder. 
The target is to reduce the loss, i.e., to replicate the original input. 
We utilize a sequence-to-sequence AE model, 
where both the encoder and the decoder are recurrent time-depended ANNs using several gated recurrent units (GRUs). 

\begin{figure}[h]
 \centering
 \includegraphics[width=0.8\linewidth]{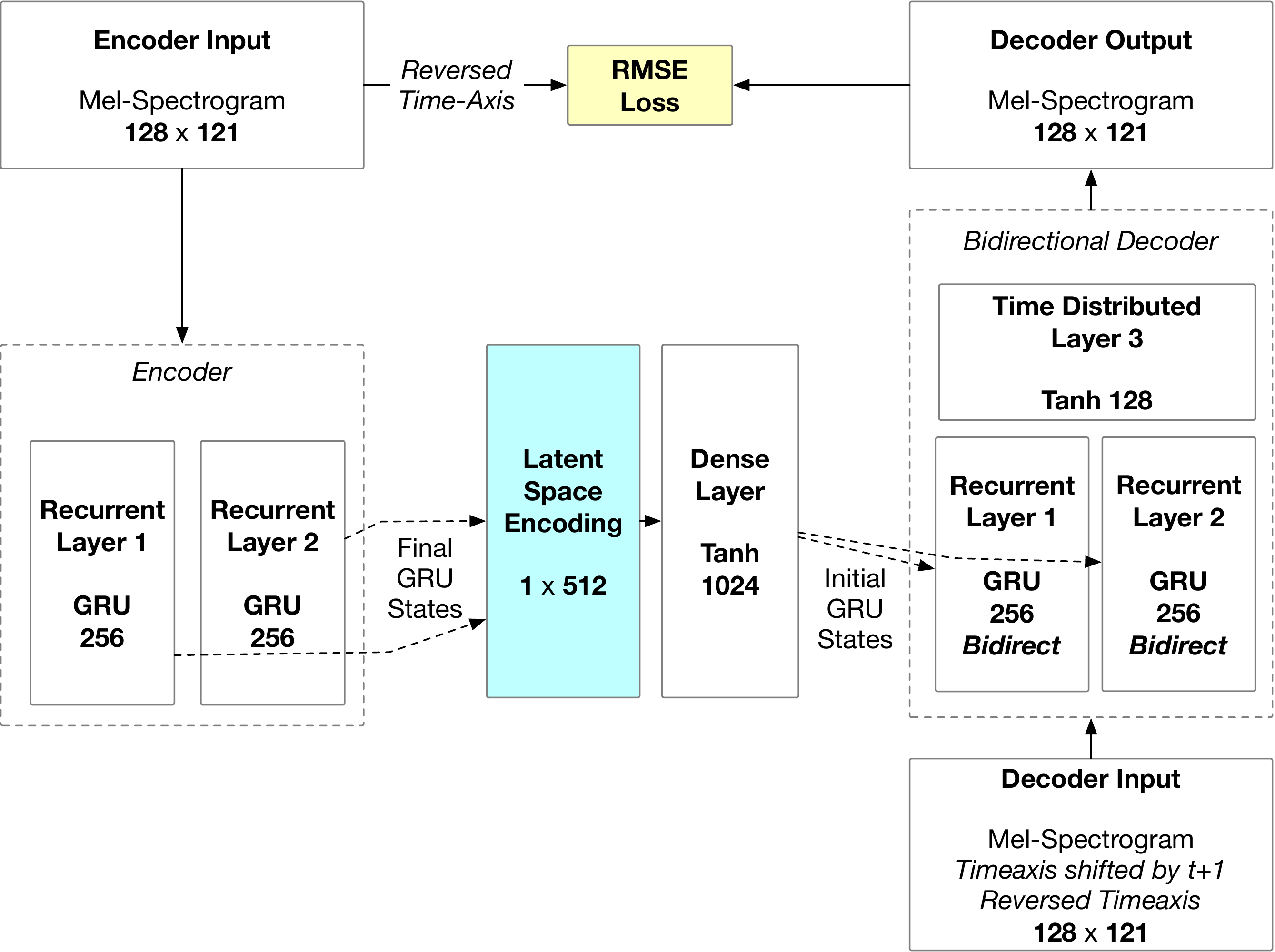}
 \caption{Sequence-to-Sequence Autoencoder structure. The Encoder and Decoder utilize a similar, recurrent structure with two layers of 256 GRUs. Moreover, the decoder utilizes bi-directional layers. The latent encoding layer is connected to both encoding layers' final hidden states, resulting in a 512-length feature vector for each Mel-spectrogram input. After transformation with a dense layer, the feature vector is used to initialize the decoder layers' hidden states. The shifted and reversed Mel-frequency are the inputs to the decoder. The model is trained using the RMSE-Loss between the reversed input and the reconstruction of the decoder.}
 \label{fig:ae}
\end{figure}

Figure~\ref{fig:ae} illustrates our AE structure, 
which closely follows the structure from AuDeep~\cite{Frei18a}, 
as used for sequence-to-sequence learning from audio~\cite{Amir17a,Suts14a}. 
The encoder and decoder consist of 2 recurrent layers, 
each with 256 GRU units and tanh activation functions. 
All layers, except the latent encoding and output, utilize a 20\% dropout. 
The final GRU cell states serve as the latent encoding with a significant dimension reduction. 
The encoder inputs are the Mel-spectrums with dimensions Mel-bins $\times$ time-steps (128$\times$121), 
while the data encoding is one-dimensional vectors, 
consisting of the final states of the GRUs in the encoder (512$\times$1). 

The decoder is set to use bidirectional GRU layers, 
using forward and backward passes, i.e., past and future information. 
This doubles the number of hidden states and lets the bidirectional decoder end up with 512 hidden states per decoder layer and a total size of 1024 states to initialize. 
The latent encodings are fed through a dense layer with an equal number of units for the initialization of the decoder hidden states. 

The decoder inputs are the Mel-spectrograms shifted by $t+1$ (with a leading zero), 
which are further reversed in the time-axis~\cite{Amir17a}. 
A time-distributed layer is connected to the second GRU layer with a fully-connected dense layer of 128 tanh units, 
equal to the number of required Mel bins to restore. 
Finally, the loss is the root-mean-square error (RMSE) between the time-axis reversed original Mel-spectra and the decoder output sequence. 

We are interested in the generation of latent data encoding, 
because it is an effective automated dimension reduction technique 
and allows employing simpler and more robust classification models. 
\subsection{Classification and Anomaly Detection}\label{ssec:networksstructures}
The reduced encoding allows us to employ a small and fast learning classification model in form of a multi-layer perceptron (MLP).  
The training of the MLP model for the audio classification requires a labeled data set.
Labels can be created in various ways:  
\begin{itemize}
\item Conducting experiments with different events to generate event observations.
\item Manually augmenting live observations if an event cause is known, e.g., a boat was in the water and observed by the user.
\item Manually augmenting observations which have been classified as anomaly.
\end{itemize}

\begin{figure}[h]
  \centering
  \includegraphics[width=0.65\linewidth]{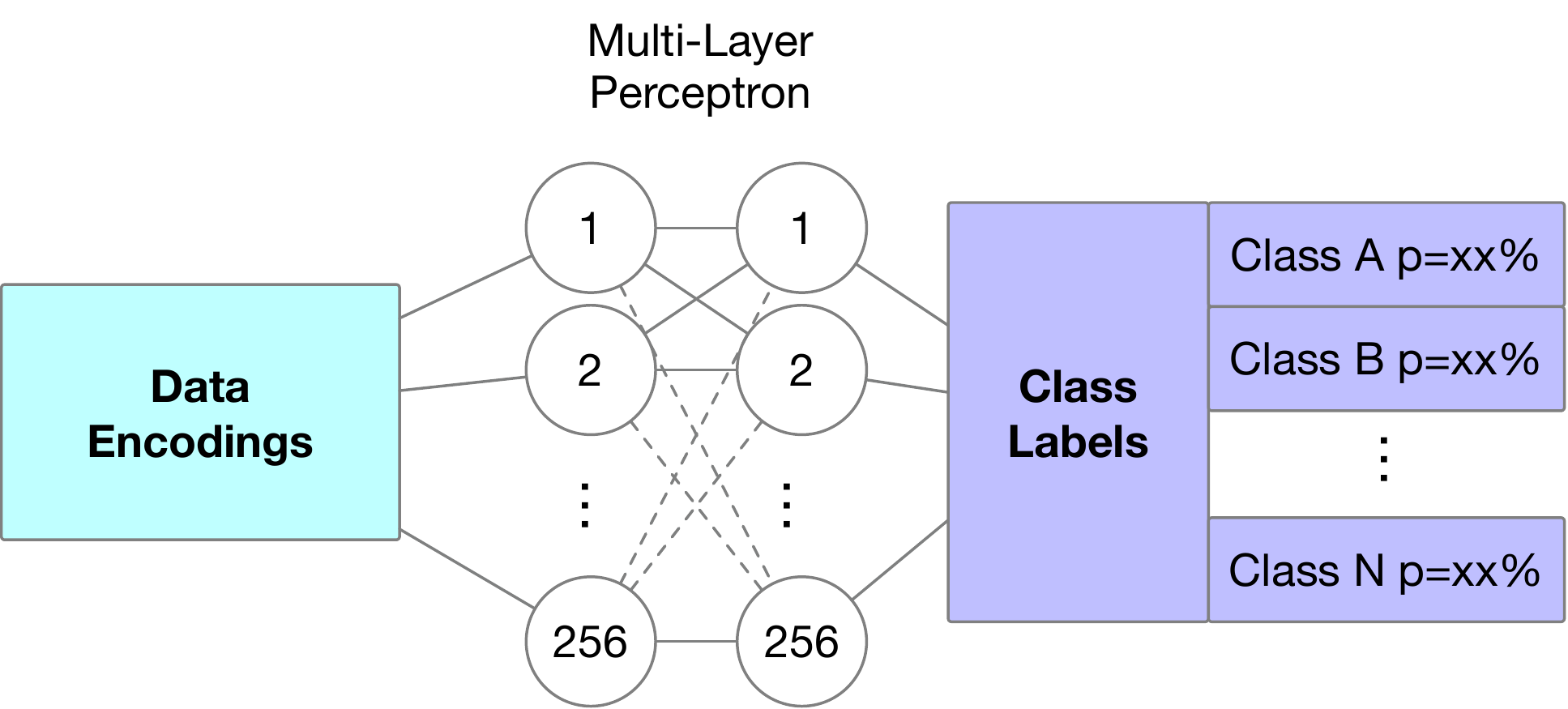}
  \caption{Multi-Layer Perceptron. The structure has two fully connected layers with 256 tanh units as activation function. The input are the AE encoded data, while the output represents the probability for an input belonging to each class.}
  \label{fig:mlp}
\end{figure}

Hence, during the live operation of our system, 
the available labeled data grows
as additional event classes are added. 
This also requires a regular retraining of all the models to include new labels.  
The MLP model utilizes a two-layer structure with 2 x 256 rectified linear units. 
The data encodings with length 512 are used as input for the MLP,
which returns a vector of class probabilities, 
as visualized in Fig.~\ref{fig:mlp}.  

The AE RMSE loss can be utilized for the detection of anomalies in the observations. 
During the training phase, the loss is minimized to the training data, 
which is in optimal case a stratified sample of the complete set of occurred events. 
If the AE is then applied to a new observation and a high reconstruction loss occurs, 
this observation is likely to being not, or very rarely, present in the training set. 
It can thus be considered an anomaly, i.e., a very unlikely happening or not yet observed event.  

\subsection{Sound Localization}

The localization of many underwater sound sources is possible using an array of hydrophones. However, depending on the type of sound source (static or dynamic) and its acoustic emissions (length and frequency content of the sound), different algorithmic approaches must be used to localize it. Most sound localization algorithms use a combination of time delays in the sound recordings of different hydrophones together with sound wave geometries (represented as sound rays) to determine the position of sound sources~\cite{Duan14a}. To enhance the precision in the localization, algorithms must take into account variations in the transmitting medium parameters such as water temperature, pressure, density, and currents~\cite{Thod04a}. The noise and the hydrophones' sensitivity are usually incorporated as part of the uncertainty of the acoustic measurements~\cite{Skar18a}.
One situation where acoustic triangulation methods cannot determine a sound source's position is when a static source is permanently emitting non-varying sound. In this situation, all the hydrophones in a network record the same acoustic signals over time, and no time delays or acoustic changes can be measured in the time domain. In the frequency domain, the constant frequency-energy content provides no additional information for the localization.
We were able to determine the location of sound sources in the water dam using a hydrophones network. Our system uses time-delay estimation and acoustic ray geometries to solve a simultaneous system of quadratic equations. Our localization experiments and results are presented in Section~\ref{sec:experiments}.

\subsection{Data Recognition Pipeline}
\begin{figure}[h]
 \centering
 \includegraphics[width=0.95\linewidth]{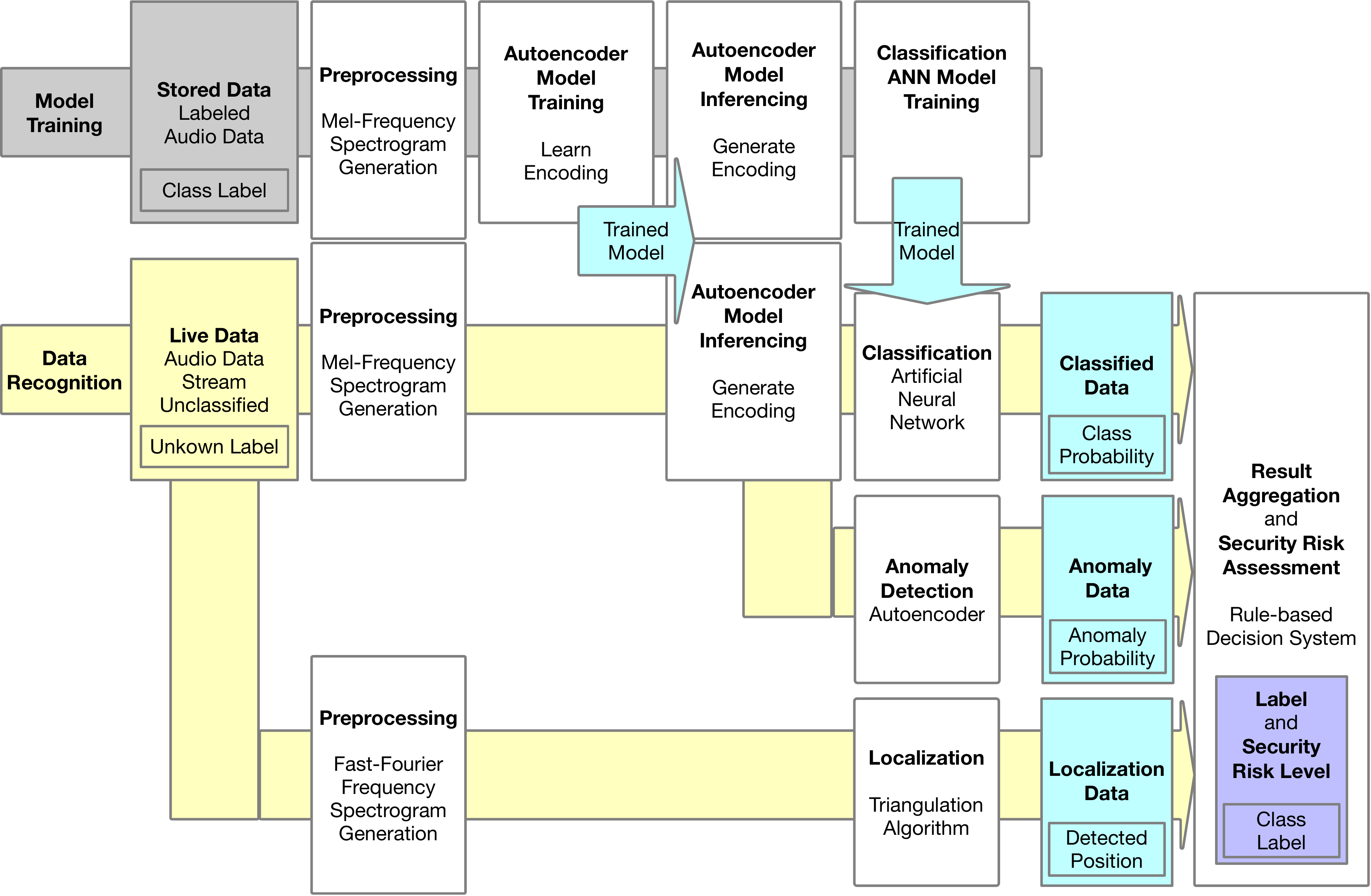}
 \caption{
Data Recognition Pipeline. 
The pipeline is divided into two parts: the upper (grey) pipeline is dedicated to the training of the AE and MLP models using the preprocessed Mel-spectrograms. 
The learned models are then applied for inferencing in the lower pipeline (yellow), dedicated to real-time data recognition and model inferencing. 
The lower pipeline utilizes different modules for classification, anomaly detection, and localization. 
The classification and anomaly detection models are based on the pre-trained models and the AE feature processing. 
The results from the three modules are further aggregated for a final security risk assessment.}
 \label{fig:pipe}
\end{figure}

The complete data recognition pipeline is outlined in Fig.~\ref{fig:pipe}. 
It is divided into model training and lives data recognition. 
With this, the model training relies on substantial amounts of stored and labeled data, 
while the live recognition ensures immediate processing of incoming data recorded by the sensor array. 
As mentioned in Section~\ref{sec:transmission}, the model training is executed on a GPU cluster at an external site, 
while the recognition is executed on-site with an industrial PC situated in the dam wall.  

The model training utilizes selected stored audio data and their event class labels. 
The training set is pre-selected to represent all present event classes in a balanced set 
to avoid overfitting to a specific event class.
The data is then preprocessed as described in~\ref{ssec:prepro} and used to train the AE model. 
The data encoding produced by the AE after training, 
together with the class labels, 
are then utilized for training the MLP for classification. 
The model training can be conducted independent of the live recognition part, 
which allows intense model tuning on the cluster. 
Both models can be trained and optimized individually, also in different intervals. 
However, it is recommended to adjust both models if the training data changes. 
High-quality models can then be transferred to the industrial PC for the live recognition of incoming events.  

The incoming live data stream consists of the live acoustic data recorded by the 5-hydrophone sensor network, 
which is segmented into 6s long observations. 
These observations are first preprocessed, 
either to time-frequency spectrograms or Mel-spectrograms. 
Then, the AE model is applied to the Mel-spectrograms to create data encodings for the incoming data stream. 
The encodings are used as input to the classification algorithm, 
which outputs a probability of the observations belonging to a particular class.  
Further, the anomaly detection creates a probability of each observation being an anomaly, and lastly, 
the localization returns an approximated position of the acoustic signal source in the water.  

Finally, the resulting probabilities are aggregated and labeled. 
Based on a parameterized decision tree, each observation is evaluated and given a security risk assessment based on the information of the classification, anomaly detection, and localization algorithm. 
This final risk assessment is discussed in Section~\ref{sec:fin}.  

\subsection{Software Platform: Cognitive Architecture}

\begin{figure}[h]
 \centering
 \includegraphics[width=0.6\linewidth]{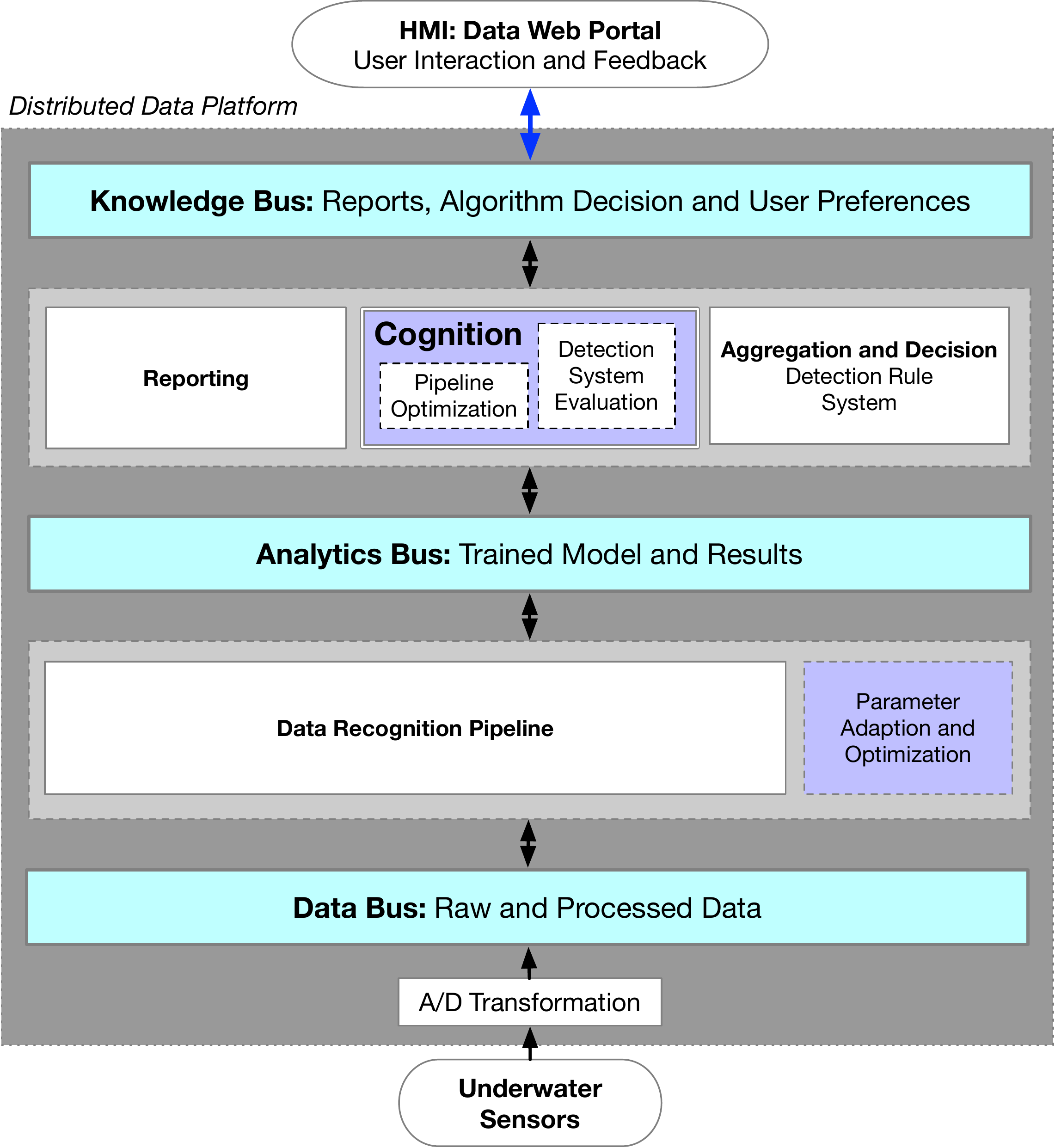}
 \caption{Cognitive Platform based on the CAAI architecture~\cite{Fisc20a}. 
 The architecture includes all pre-processing steps, from gathering the data to the visualization and user feedback. A central part of the processing is the data recognition pipeline, which is extended by a module to adapt and optimize the connected algorithms and models' parameters. The cognitive part is implemented as a method for automatic tuning and model selection to ensure optimal parameter settings during training and inferencing in cohesion with the desired user targets. }
 \label{fig:caai}
\end{figure}

The AI solution's architecture for detecting security threats is based on the cognitive architecture for artificial intelligence in cyber-physical systems (CAAI)~\cite{Fisc20a}. 
It includes an explicit interface to the user (HMI), 
where interaction between user and processing pipeline is possible, 
i.e., changing goals or inserting information, 
which the software can request from the user in the web portal. 
The complete architecture is displayed in Fig.~\ref{fig:caai}. 
The main elements of the platform are three dedicated information buses and several modules summarized into two layers: 
Firstly, the data recognition pipeline, 
which is extended by a module for parameterization and optimization, 
and secondly, the conceptual layer where the reporting, the cognition, 
and rule-based decision modules are situated. 
This architecture is considered cognitive~\cite{Fisc20a} 
as it includes methods to evaluate and optimize the processing modules automatically, 
i.e., the data recognition pipeline. 
In our case, the parameters of the underlying algorithms are adapted to ensure a high ML model quality 
by establishing a low training and test error. 
Several parameters are optimized.
Examples are parameters controlling the network topology, i.e., number of layers, number of neurons, 
of the Autoencoder and Perceptron. 
Further examples are training parameters, such as selecting an optimizer, learning rate, or batch size.

\subsection{User Interaction}
The classification of audio data into predefined classes requires labeled audio data. 
That is, an expert or user needs to define whether specific audio signals belong to certain classes. 

The problem is that data labeling can be extremely time-consuming. 
A user might have to listen to many hours or even days of audio data. 
This is clearly not feasible.
Moreover, while the user may know that a particular event occurred 
(e.g., from human reports and observations that were made on-site), 
it would still be challenging to find the corresponding set of audio signals for that event and assign labels 
(e.g., because of vague time information). 

A solution to this challenge is to use the unsupervised part of the detection process as a pre-filter (anomaly detection). 
A user should only have to screen those signal samples 
that are determined to be relevant or anomalous by the unsupervised learning models.  

To enable this exchange between user and algorithms, an interface is required. 
This interface allows the user to screen the various audio samples (based on the unsupervised analysis). 
Once the user has identified certain classes of events (e.g., sounds of a diver or sounds of a boat), 
the interface has to collect that information and store it in the database. 
Subsequent training runs of the classification algorithm will then integrate those new labels into the learning process. 

\begin{figure}[h]
  \centering
  \includegraphics[width=0.7\linewidth]{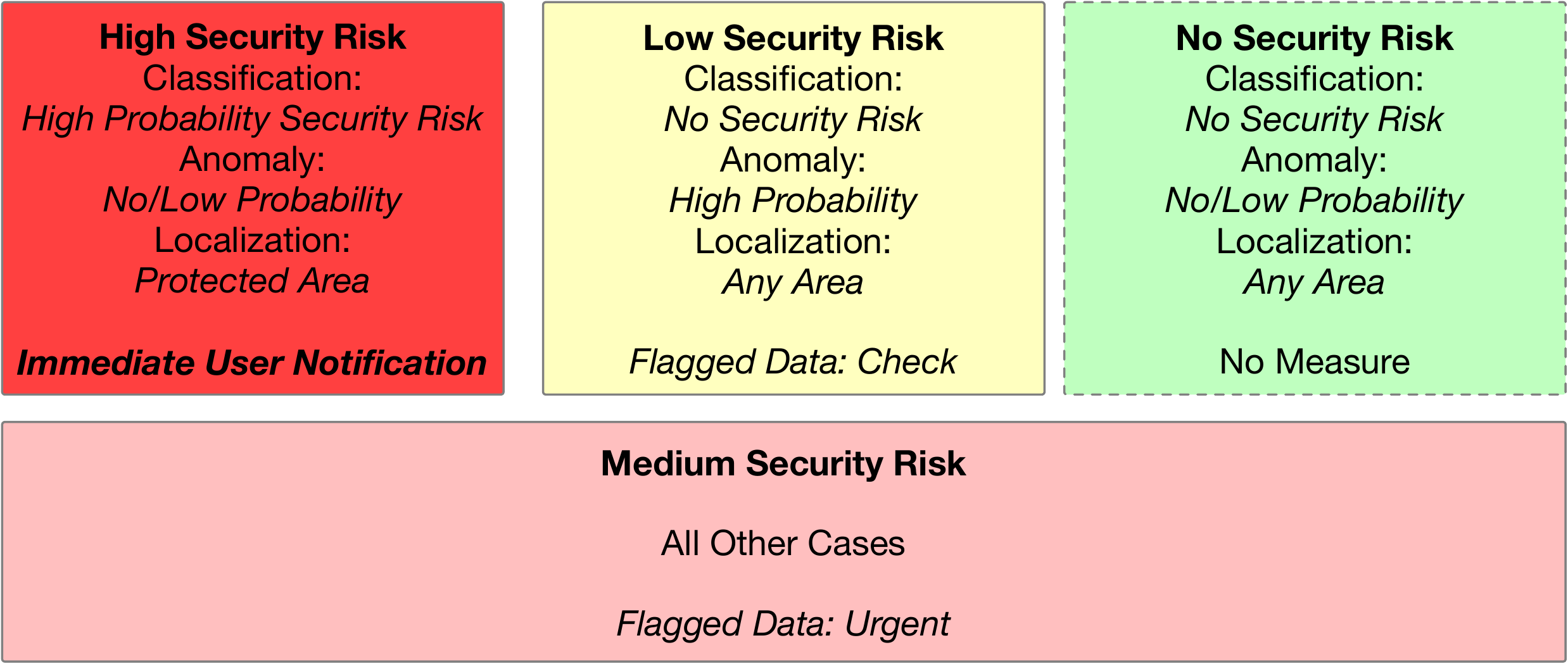}
  \caption{Security Risk Level Assessment. Based on each processing module, the final risk assessment is conducted based on a rule-based system, which finally leads to different user notifications.}
  \label{fig:riskLevels}
\end{figure}

Moreover, the user can directly influence the rule-based decision system's parameters, 
which is connected to the detection pipeline and aggregates the results to a final label. 
As visualized in Fig. \ref{fig:riskLevels}, the final risk assessment is based on the user preference, and the rule-based system can either prioritize the classification, anomaly detection, or localization. 
The user is presented with the results and can interactively shift the parameters to determine a specific setup's impact.  

\section{Experiment Case Studies}\label{sec:experiments}
\subsection{Dataset}

The experiment Dataset is a composition from field tests and laboratory testing. 
The laboratory testing took place utilizing a large swimming pool in a lab environment, 
normally used for testing underwater-robots. 
Eight different noise classes were produced using simple objects, 
such as wooden planks, tubes and cases. 
In experiments, with these objects a set of actions were performed underwater and recorded by a hydrophone: 
Knocking on wood/plastic/concrete wall (3 classes), 
emerging small and large bubbles (2 classes), 
metal clanking (1 class), 
plastic scratching (1 class), 
and plastic knocking and scratching (1 class). 

The recorded field test set includes only two classes which were aggregated and anonymized and include:  
\begin{itemize}
\item Observations where no events occurred: Considered as “normal” operating state and identified as normal environmental noise  
\item Observations where high-risk events occurred: Considered to trigger an immediate user notification and identified as high-risk danger 
\end{itemize}

The recorded sounds were split into samples of 6 seconds for a total of 1399 sound samples. 
Further, these samples were preprocessed to Mel-Spectrograms as described in Section~\ref{ssec:prepro}.     

\subsection{Case Study: Classification with Autoencoder and Multi-Layer Perceptron }
As a first case study, 
we try to learn a classifier that distinguishes the ten classes described above. 
We use the Autoencoder
to generate the latent encoding and then train an MLP on the latent encoding described in section~\ref{ssec:networksstructures}. 
Both models are implemented in Keras / Tensorflow~\cite{Chol15a,Abad15a}.  

The AE is trained for 250 epochs using a batch size of 96. 
Weights are optimized with ADAM~\cite{King15a}, and a learning rate of 0.001 is employed. 
The MLP is trained for 200 epochs with a batch size of 96 and a learning rate of 0.001 using RMSprob.  
The training of both AE and MLP is performed only with the training data. 
Afterwards, the classes are predicted by the composite Model (AE\&MLP) based on the test data. 
The training and test procedure are repeated for ten different random splits of the training and test data.  

To judge the quality of results, we perform an additional experiment that serves as a baseline. 
For this experiment, the classes are predicted by a simple nearest-neighbor model. 
That is, the class of any test sample is predicted to be the class of the training sample that has a minimal distance to the test sample. 
The distances are computed directly on the preprocessed input data (i.e., the Mel-spectrograms). 
Hence, our nearest-neighbor baseline is independent of the AE model. 

The individual classes are fairly unbalanced 
(varying between a few tenth to several thousands of samples per class). 
Therefore, we record two quality indicators: the default accuracy as well as balanced accuracy. 
The default (i.e., the usual) accuracy measure simply counts the number of correctly predicted classes. 
If one class outnumbers the other classes by a large margin, 
even a model with very poor performance might receive relatively large quality values simply by always predicting the most frequent class. 
The balanced accuracy attempts to avoid this issue, 
by first computing the fraction of correctly identified samples for each singles class,
then summarizing these statistics by taking the mean over all classes. 

\begin{figure}[h]
 \centering
 \includegraphics[width=\linewidth]{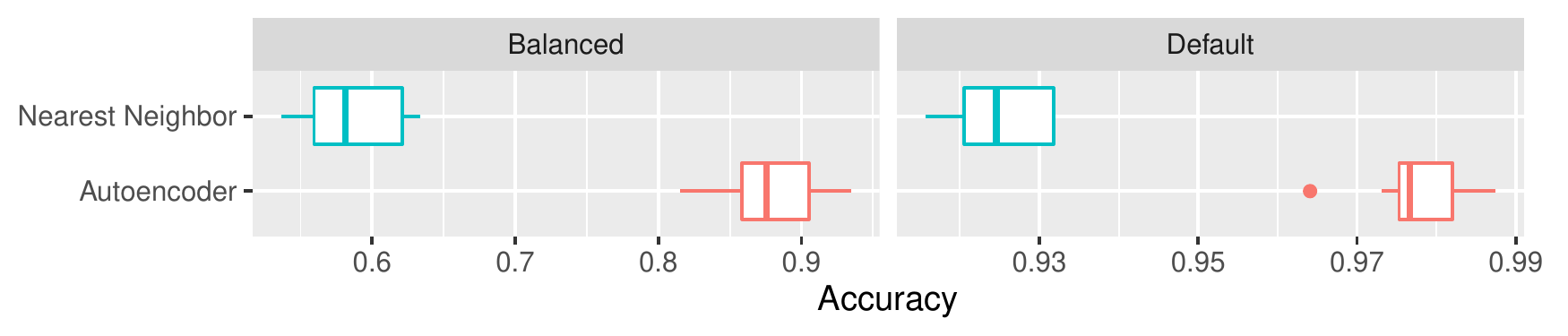}
 \caption{
 Visualization of the results from the classification case study. 
 The left facet shows balanced accuracy, 
 the right facet shows the default accuracy. 
 Each of the boxes represents 10 model evaluations (one evaluation for each data split).
 }\label{fig:acc}
\end{figure}

The corresponding values of the default and balanced accuracy, 
which were observed in the experiments, 
are visualized in Fig.~\ref{fig:acc}.   
The composite AE\&MLP model outperforms the nearest neighbor baseline by a significant margin, 
regardless of the quality measure. 
The steep difference between both measures indicates that the imbalance of our data-set indeed may lead to overestimating the performance of a classifier. 
Still, with a balanced accuracy of roughly 90 \%, the resulting model seems promising. 

\begin{figure}[h]
 \centering
 \includegraphics[width=\linewidth]{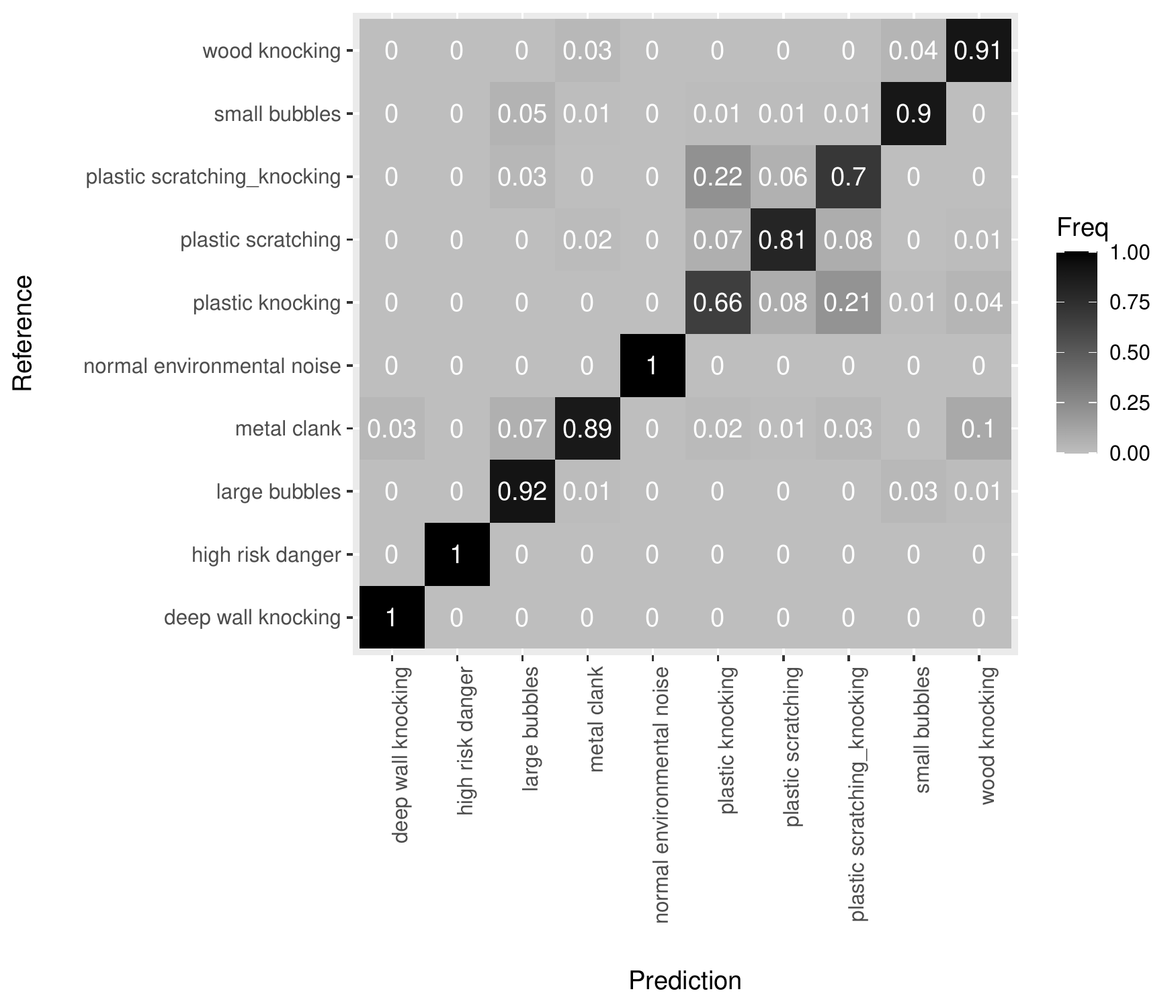}
 \caption{
 Aggregated classification results by Reference class (true class) 
 and predicted (class) over 10 repeated runs of the combined AE\&MLP classification model. 
 The numbers in each tile (also represented by tile-color) 
 represent how frequently the respective result was observed on the test data set. 
 }\label{fig:confusion}
\end{figure}

As a next step, it is, of course, of interest to observe which individual classes are predicted well, 
and which classes are not. 
For that purpose, we show an aggregation of the confusion matrices 
computed for each of the 10 runs in Fig.~\ref{fig:confusion} 
(only based on the AE\&MLP model). 
Each of the tiles shows the corresponding frequency of the results, 
averaged over the ten runs. 
Thus, the numbers on the dark-shaded diagonal indicate the respective accuracy values for each class. 
Off-diagonal values indicate the frequency of respective misclassification cases.  

The figure shows that the worst results are observed for the classes 
``plastic knocking'', ``plastic scratching'', and ``plastic scratching\_knocking''. 
The most frequent error is a mix-up of 
``plastic knocking'' and ``plastic scratching\_knocking''. 
A likely explanation is the inherent similarity of these classes. 

The best performance is achieved for 
``deep wall knocking'', ``high risk danger'', and ``normal environmental noise''. 
A likely explanation for ``high risk danger'', 
and ``normal environmental noise'' 
is the respective number of samples in those classes,
as they are the most frequent classes in our data set. 
The class ``deep wall knocking'' on the other hand, seems to be fairly different from all other classes, 
rendering it easy to identify. 

\subsection{Case Study: Anomaly Detection}
The anomaly detection case study generally uses the same software and data set. 
However, some significant differences have to be mentioned. 

To provide a reasonable test case, 
we generate anomalies by removing all samples of a single class from the training data sets. 
We denote the removed class as the holdout class. 
All samples of the holdout class are included in the test set. 
When predicting the test set with the AE model, 
all samples of the holdout class will be treated as true anomalies, 
whereas all other test samples are treated as non-anomalous data. 
To investigate this result's dependency on the holdout class, 
we perform repeated experiments where different classes are treated as a holdout class. 
We limit this to the eight minority classes. 
The two more frequent classes 
(``high risk danger'', ``normal environmental noise'') 
are not used as holdout classes. 

Otherwise, the AE network structure and training remains largely unchanged compared to the classification test study 
(250 epochs, batch size of 96, ADAM with learning rate 0.001). 
Unlike with classification, the test evaluation does not employ the latent space. 
Instead, the autoencoder's decoder network is used to compute the reconstructed data for each test sample.
Then, the error for each test sample 
(based on the difference between actual and reconstructed data) 
is computed.  

Afterwards, the actual anomaly detection is performed by comparing the reconstruction error from each test sample to a threshold. 
More significant errors lead to an identification of a test sample as an anomaly. 
Just as in the classification case study, 
we need a baseline to judge how well our AE model performs. 
Again, we use a simple nearest-neighbor model. 
For each test sample, we compute the distance to the nearest neighbor in the training data set (on Mel-spectrograms). 
If that distance becomes large (i.e., larger than some threshold), 
the corresponding sample is identified as an anomaly. 

Clearly, deployment of this model in the real world would require 
to set a specific threshold for each case. 
The threshold will affect how many anomalous samples are recognized correctly (true positives), 
and how many false positives are generated. 
The correct trade-off between true and false positives will depend on the application 
(e.g., how many samples can be reviewed by a human operator). 
This cannot yet be determined clearly for the data set under investigation.  

Hence, we evaluate our model (and the baseline) 
without a fixed threshold and compute the AUC, 
the Area Under the Receiver Operator Characteristic (ROC) Curve. 
The ROC curve represents all possible trade-offs between true positive rate and false positive rate 
(each individual trade-off represents one threshold value). 
A random guess would result in an AUC of 0.5, 
whereas an ideal model would achieve an AUC of 1.0. 

\begin{figure}[h]
 \centering
 \includegraphics[width=\linewidth]{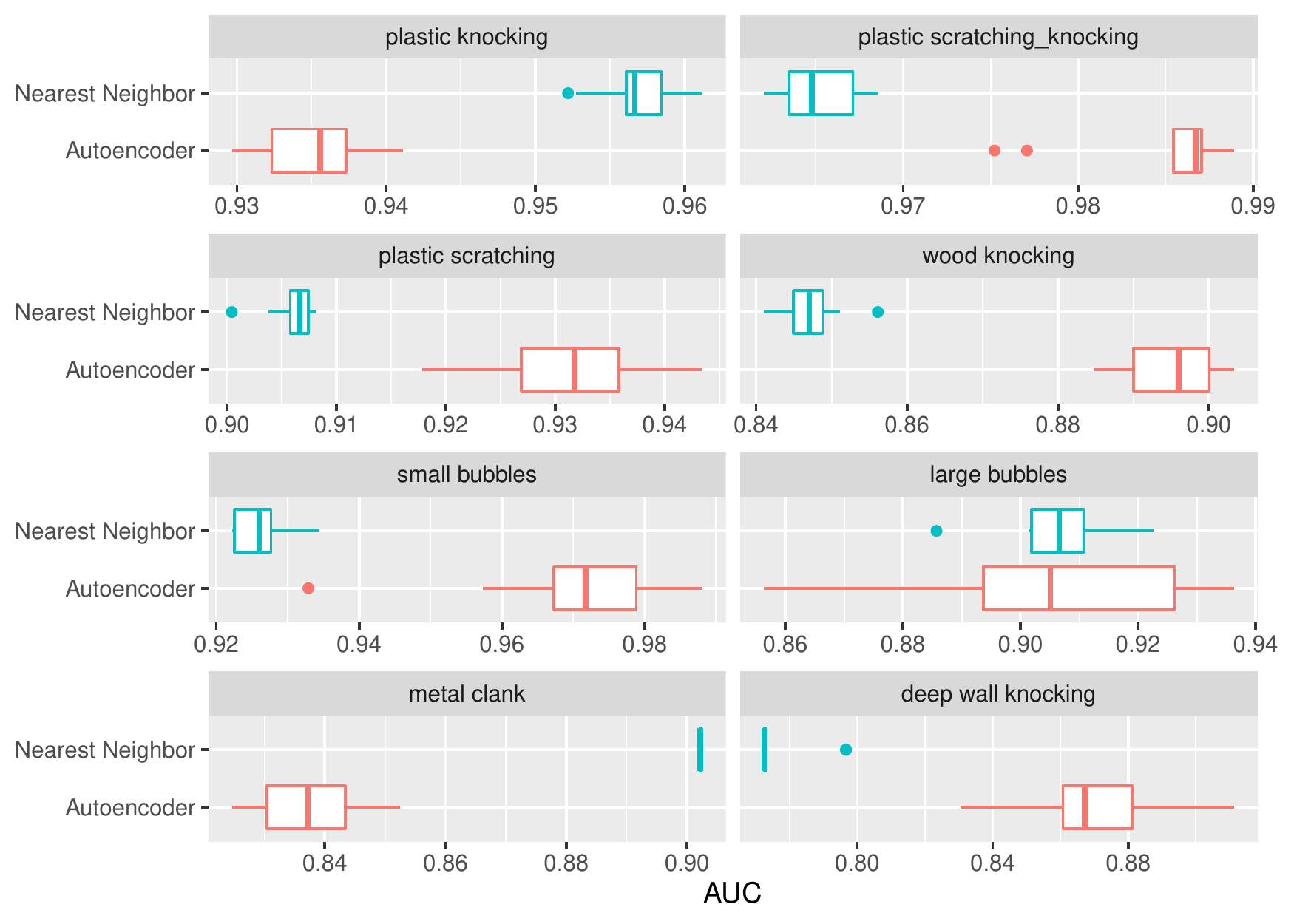}
 \caption{Boxplot of the Area Under the ROC Curve (AUC) for each holdout class for the anomaly detection case study.}
 \label{fig:auc}
\end{figure}

The AUC values for each holdout class are reported in Fig.~\ref{fig:auc}. 
The AE outperforms the baseline for five of the eight holdout classes. 
For the holdout class ``large bubbles'', 
no clear difference can be observed 
(although the AE seems to produce a more considerable variance). 
For two of the five classes, the baseline outperforms the AE:
``plastic knocking'' and ``metal clank''. 
The reasons for this latter observation are not entirely clear, 
but it may hint at a more extensive intra-class similarity in the Mel-spectrogram data, 
which would be favorable for the nearest neighbor model. 

\subsection{Case Study: Localization of Underwater Sound Sources Using Passive Acoustic Monitoring with a Hydrophone Network}

\begin{figure}[h]
  \centering
  \includegraphics[width=\linewidth]{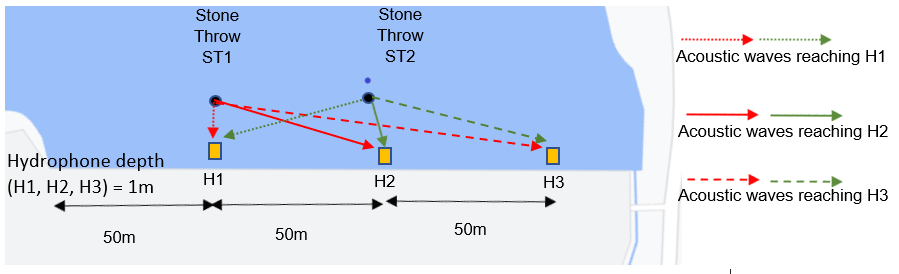}
  \caption{Localization Experiment. The 2D position of stones entering the water (ST1, ST2) is determined from the time delays in the sound recordings of hydrophones H1, H2, H3. The time delays are analyzed to determine the ray geometry of the acoustic waves.}
  \label{fig:L2}
\end{figure}

We carried out localization experiments to calculate the 2D surface entry position of objects that are thrown into the water dam. For these experiments, three hydrophones H1, H2, H3, were connected to the audio amplifier and placed along the dam wall with a separation of 50m from one another. We then threw stones into the water from different locations and registered their approximate, observed water-entry locations. Our system automatically calculated the stones' water-entry position based on a time-delay analysis of the hydrophone recordings and the resulting ray geometry of the acoustic waves. Figure~\ref{fig:L2} shows an example of the configuration we used to carry out the experiments and the ray depiction of the acoustic waves that were used for calculating the specific geometries of stone throws ST1 and ST2.

\begin{figure}[h]
  \centering
 \includegraphics[scale = 0.3]{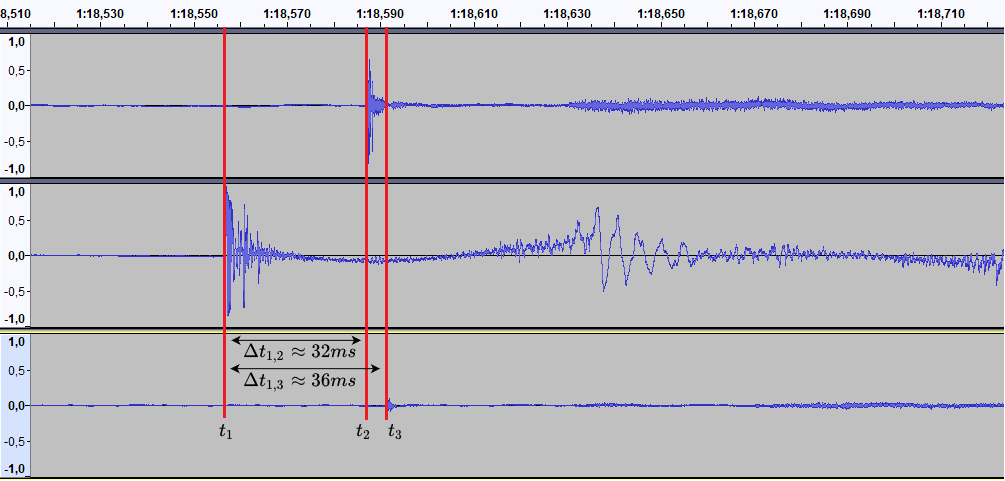}
  \caption{Sound Recordings of H1, H2 and H3 following ST2. H2 records ST2 first, suggesting that ST2 landed in the vincinity of H2. ST2 is recorded by H1 32ms later and by H3 36ms later, suggesting that ST2 entered the water between H1 and H2 and closer to H2.}
  \label{fig:L5}
\end{figure}

Figure~\ref{fig:L5} shows the recordings of H1, H2, and H3 following ST2. The uppermost acoustic wave represents the recording of H1, and the lowermost acoustic wave the recording of H3. H2 records ST2 first, suggesting that ST2 landed in the vicinity of H2. ST2 is recorded by H1 32ms after reaching H2 and by H3 36ms after reaching H2, suggesting that ST2 entered the water between H1 and H2 and closer to H2.

\begin{figure}[h]
  \centering
 \includegraphics[scale = 0.6]{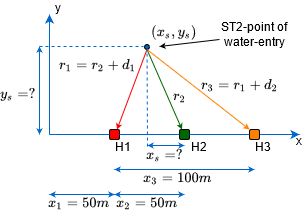}
  \caption{The acoustic rays used to calculate the entry point ($x_s, y_s$) of ST2 into the dam. $d_1$ and $d_2$ are distances that can be determined from the time-delays observed in the hydrophone recordings, $r_1$, $r_2$ and $r_3$ are the distances from ST2 to the hydrophones. }
  \label{fig:L6}
\end{figure}

The hydrophones were placed at a shallow depth of 1m so that the 3D geometry analysis of the acoustic rays can be simplified to a 2D scene by disregarding the hydrophone depth and assuming the hydrophones are close to the water surface. The system first analyses the time delays in the acoustic recordings to determine the region where ST2 entered the water, in this case, between H1 and H2 and closer to H2. A 2D geometry is built using the region information from which the water-entry point ($x_s, y_s$) of ST2 can be calculated. The 2D-geometry used for localizing ST2 is shown in Fig.~\ref{fig:L6} and is the same 2D geometry used for all stone throws entering the water in the region between H1 and H2 and closer to H2. 

\begin{figure}[h]
  \centering
 \includegraphics[scale = 0.2]{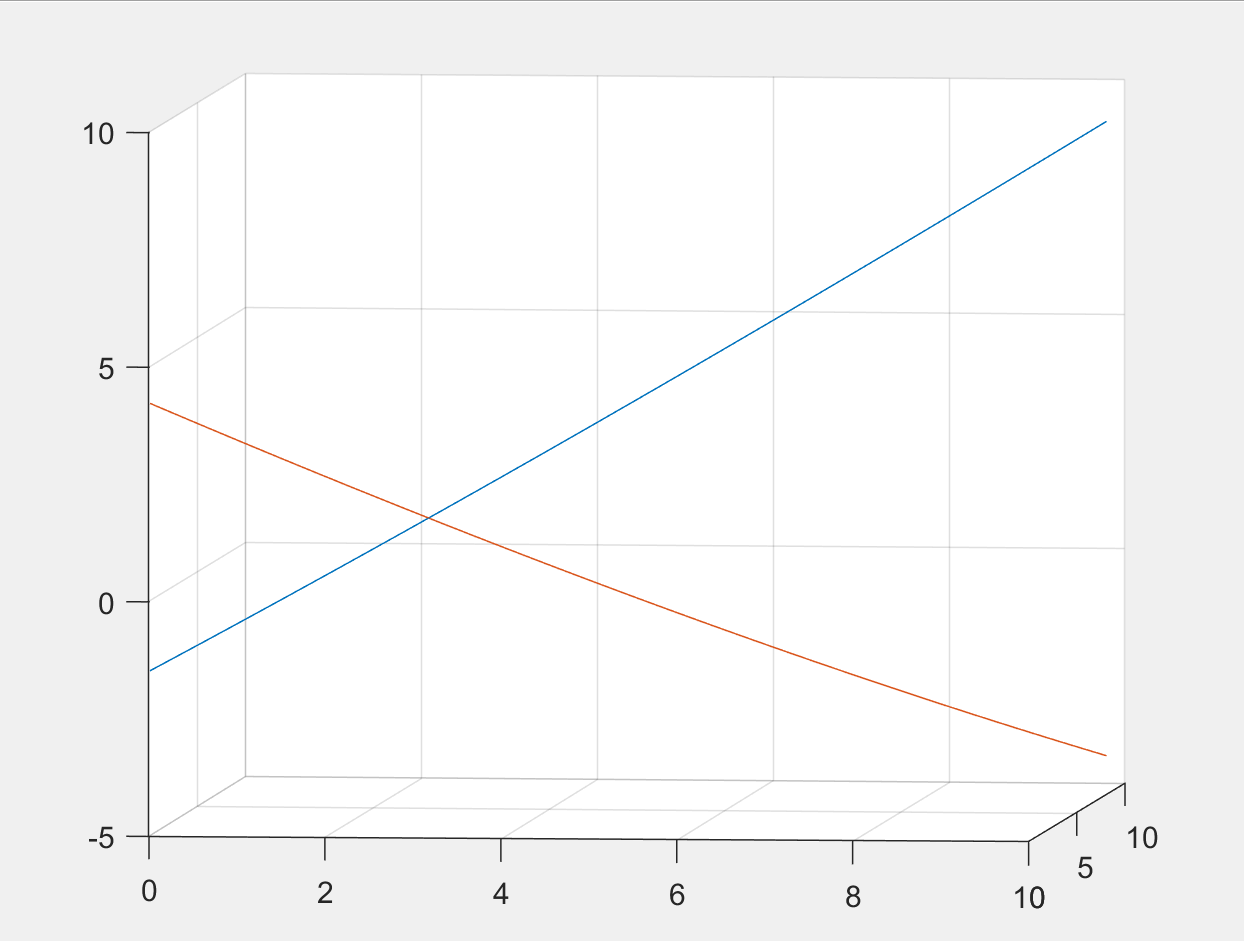}
  \caption{Results of iteratively solving the system of simltaneous quadratic equations for the 2D acoustic geometry of ST2 over a range of 10m for  ($x_s, y_s$). The obtained result  ($x_s, y_s$) = (2.8, 1.5) precisely matches the experimental observation of the location of ST2.}
  \label{fig:MinSquare2}
\end{figure}

From the geometry in Fig.~\ref{fig:L6} we can build the following system of simultaneous quadratic equations:
\begin{equation}
r_2^2 = x_s^2 + y_s^2
 \label{eq:L1}
\end{equation}	
\begin{equation}									
r_1^2 = (r_2 + d_1)^2 = (50 - x_s)^2 + y_s^2 	
 \label{eq:L2}
\end{equation}
\begin{equation}							
r_3^2 = (r_1 + d_2)^2 = (r_2  + d_1 + d_2 )^2 = (50 + x_s)^2 + y_s^2
 \label{eq:L3} 
\end{equation}						

to determine $r_2, x_s$ and $y_s$. $r_2$ is an unknown distance because, from the experiment, we cannot measure how long it took for the acoustic waves of ST2 to reach H2, we can only measure the time delay that the acoustic waves had in reaching H1 and H3 after reaching H2.
Using v = 1430 m/s for the speed of sound in water (considering the experiments were carried out in winter by near freezing temperatures), $\Delta t_{H2H1} = 32ms$ and $\Delta t_{H1H3} = 36ms$, gives the result $d_1 = v * \Delta t_{H2H1} = 45.76m$ and  $d_2 = v *  \Delta t_{H1H3} = 51.48m$. 
Eqs.~\ref{eq:L1} -  ~\ref{eq:L3} can be solved iteratively for all feasible values of  ($x_s, y_s$). Figure~\ref{fig:MinSquare2} shows the results of the iterative solution using a solution range of 10m for $x_s$ and 10m for $y_s$. As can be seen, the results obtained  ($x_s, y_s$) = (2.8, 1.5) match the experimental observations with very good precision. 

\begin{figure}[h]
  \centering
 \includegraphics[scale = 0.3]{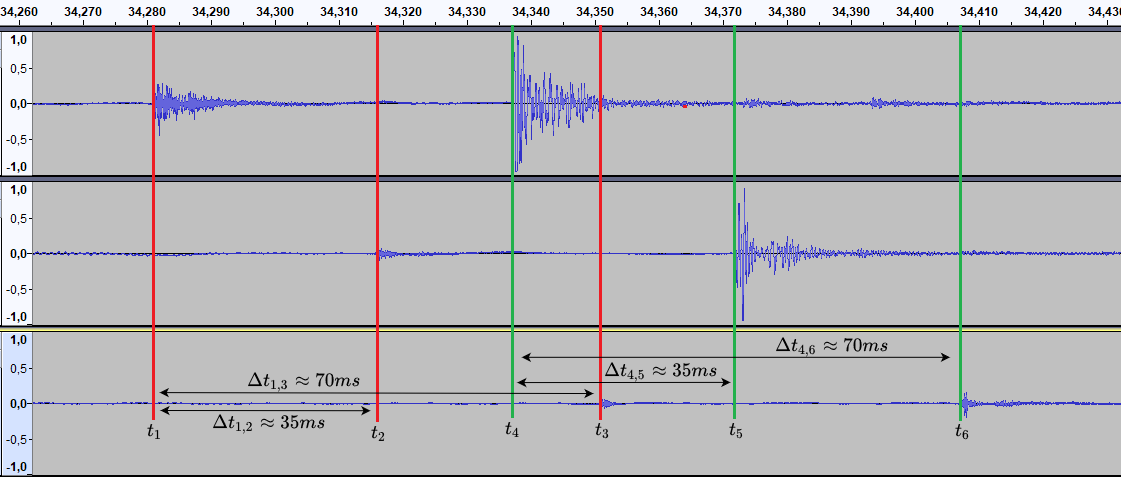}
  \caption{Sound recordings of H1, H2 and H3 following ST1. The uppermost acoustic wave represents the recording of H1 and the lowermost acoustic wave the recording of H3. H1 records ST1 approx. 35ms earlier than H2 and 70ms earlier than H3, indicating that ST1 entered the water very close to H1.}
  \label{fig:L3}
\end{figure}

Figure~\ref{fig:L3} shows the recordings of H1, H2, and H3 following ST1. The uppermost acoustic wave represents the recording of H1, and the lowermost acoustic wave the recording of H3. The acoustic waves produced by ST1 are first recorded by H1, indicating that ST1 entered the water in the vicinity of H1. The acoustic waves of ST1 reached H2 with a delay of 35ms after reaching H1 and H3 with a delay of 35ms after reaching H2, giving further indication that ST1 entered the water very close to H1. During the ST1 experiment, the stone broke into two parts, the acoustic waves of which (first a quieter, smaller stone and then a louder, larger stone) can both be seen in the acoustic recordings of H1, H2, and H3. 

\begin{figure}[h]
  \centering
 \includegraphics[scale = 0.6]{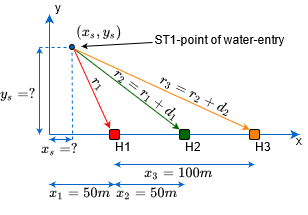}
  \caption{The acoustic rays and time delays in hydrophone recordings used to calculate the point of entry  ($x_s, y_s$) of ST1 into the dam. $d_1$ and $d_2$ are distances that can be determined from the time-delays observed in the hydrophone recordings, $r_1, r_2$ and $r_3$ are the distances from ST1 to the hydrophones.}
  \label{fig:L4}
\end{figure}

The 2D-geometry used for localizing ST1 is shown in Fig.~\ref{fig:L4} and is the same 2D geometry used for all stone throws entering the water in the vincinity of H1. 

\begin{figure}[h]
  \centering
 \includegraphics[scale = 0.2]{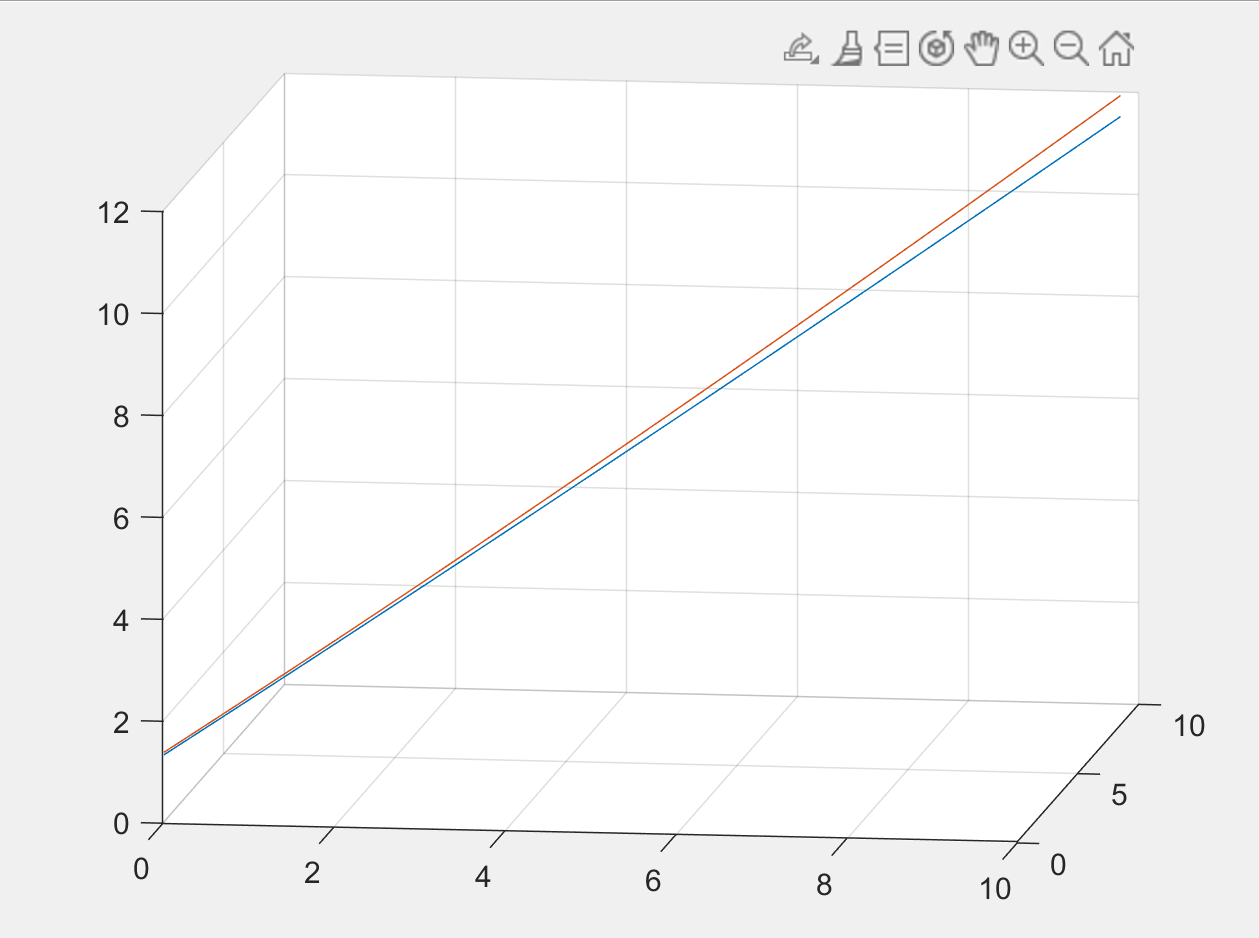}
  \caption{Results of iteratively solving the system of simltaneous quadratic equations for the 2D acoustic geometry of ST2 over a range of 10m for  ($x_s, y_s$). The obtained result ($x_s, y_s$) = (2.8, 1.5) precisely matches the experimental observation of the location of ST2.}
  \label{fig:MinSquare1}
\end{figure}

Using v = 1430 m/s for the speed of sound in water, $\Delta t_{H1H2} = 35ms$ and $\Delta t_{H2H3} = 35ms$, gives the result $d_1 = v * \Delta t_{H2H1} = 50.05m$ and  $d_2 = v *  \Delta t_{H1H3} = 50.05m$. Solving iteratively for eqs.~\ref{eq:L1} -  ~\ref{eq:L3} with 10m solution ranges for $x_s$ and for $y_s$ gives the results shown in Fig.~\ref{fig:MinSquare1}. The results obtained  ($x_s, y_s$) = ($\approx$ 0.0, 1.5) also match the experimental observations. 

From the two experiments presented in this case study, it can be seen that an array of three hydrophones placed at a shallow depth allows the calculation of the 2D coordinates of an object entering the water surface. Even though our simple experiments do not incorporate the uncertainties in hydrophone measurements, we can get reliable localization information. For sound sources emanating sound underwater (vs. at the surface of the water), the full 3D localization coordinates can be determined using a network of four hydrophones placed at arbitrary depths. For more general 3D experiments where measuring uncertainties (for example, due to water currents moving the hydrophones) and acoustic artifacts are present, we will need to model out time-delay equations probabilistically to account for measurement imprecisions and acoustic noise. 

\section{Discussion}\label{sec:discuss}
The performed classification and anomaly detection use-cases illustrate the capabilities of the AI detection system for underwater surveillance. The classification test results demonstrate an already high ability to identify events correctly based on the recorded acoustic data if precisely labeled training data is present. However, at the beginning of the system use, labeled data will not be available sufficiently. 
It is challenging to create acoustic recordings of possible risk causes based only on planned experiments. They might not reflect the authentic waterborne sounds emitted by potentially high-risk actors.
These need to be observed, detected and labeled in the real-world scenario. 
To be able to find and expose these new event classes, we employed the anomaly detection algorithm. The results indicate a high possibility of detecting noise types not present in the training sets. 
Over time, the manual checking observations labeled as anomalies by experts construct a valuable set of tagged class and training information. 
The resulting set should also be valid and transferable to other water dam locations and are thus immensely valuable for all underwater surveillance applications. 
A critical issue for all problems is the parameter setup. Machine learning algorithms are extremely parameter sensitive, and parameter choice has a substantial effect on their performance. We thus propose using a cognitive system architecture, which allows the automatic tuning of the algorithms. 
The need for such a system is visible by the anomaly detection results. 
We reported AUC values, which leave a final decision for threshold settings open. 
These thresholds directly affect how many samples will be considered as an anomaly or regular event. The choice has several implications; if set too sensitive, experts' workload to label the incoming data will be intense. On the contrary, a low sensitivity might miss essential risks, which would be a severe issue. 
However, the illustrated case study experiments only considered single observations. We will have constant surveillance and time-series data in a real-world scenario, which leads to a higher probability of correct classifications and anomaly detection. 

\section{Conclusion and Outlook}\label{sec:fin}
Due to their complexity, the safe and secure operation of water reservoirs, which belong to the critical infrastructure, is challenging. This study focused on underwater attacks. The applicability of an innovative AI-based approach for the supervision of underwater activities inside water reservoirs is demonstrated and evaluated in a real-world setting.  The implemented prototype consists of several underwater sensor networks combined with hydrophones, sonars, cameras, and drones. The hydrophone sensor network was presented in this paper. These networks are integrated into a cognitive AI architecture.
Summarizing the results from the previous sections, the paper delivers the following four main contributions:

The first significant contribution addresses technical implementation issues and is of great relevance for practitioners. The installation of such a system in the field requires the consideration of challenging physical constraints. A thoughtful description of all relevant technical aspects, essential implementation details, and problem solutions is presented. Valuable hints for practitioners are presented to avoid pitfalls and problems, e.g., acoustic interactions caused by undesired superimposed traffic noise, as well as known problems such as prevention of vandalism, maintainability, tidal variations, and many more. 
The second contribution is related to data acquisition, safe and secure transmission, storage, and data curation techniques. 
Thirdly, as the main contribution, an AI-based software solution for underwater events' recognition and location is detailed. A modular AI pipeline is implemented. It demonstrates how to provide solutions to the following relevant tasks:
\begin{itemize}
\item recognition and security assessment of unknown signals based on an unsupervised learning approach and 
\item classification of known events, which is based on a supervised learning approach.
\end{itemize}
Besides classification, the implemented AI pipeline allows the localization of events. 
The data preprocessing step, a mandatory prerequisite before AI algorithms can be applied, is described in detail. An AE approach is used to generate sequence encodings. The proposed sophisticated AE data encoding procedure allows applying a fast-learning neural network (MLP) for classification. Combining these AI pipeline modules results in transparent and well-defined AI architecture referred to as a cognitive platform because it can automatically evaluate and optimize ("tune") the processing modules.
Although the cognitive platform can be run fully automatically, some situations can benefit from user interactions. The implemented platform allows the user to screen signals and manually add further elements to the database to supervise and accelerate the learning process.

Fourthly, and finally, to demonstrate the applicability and usability of the proposed AI platform, extensive experiments were performed. The experiments combined data from lab and field data. Sound samples from several noise classes were generated. As a baseline for the comparisons with the provided AI platform, an established nearest neighbor algorithm is used. Results from three case studies (classification, anomaly detection, and localization) demonstrate the proposed approach's applicability.

In summary, a well-tested and workable prototype is available. The available results allow a generic definition of the next steps in this research project, which can be described as follows:
The transferability of our findings to other locations (water reservoirs) is of great interest. Obviously, a 1:1 copy cannot be used because every water reservoir has specific conditions (geometry of the dam, size of the reservoir, surrounding). However, the proposed approach's key components can be easily transferred to other settings, e.g., the AI algorithms can be easily adapted and trained in different locations. Furthermore, an extended comparison with additional (AI and non-AI) algorithms is of great interest.
The inclusion of results from a long-term field test, which will start soon on the actual reservoir, into our analysis is of great interest and the consideration of feedback from practitioners. The collection of data from the real-world operation will provide valuable and extensive insights into the proposed approach's working mechanisms.


\bibliographystyle{ACM-Reference-Format}
\bibliography{article}


\begin{thebibliography}{21}


\ifx \showCODEN    \undefined \def \showCODEN     #1{\unskip}     \fi
\ifx \showDOI      \undefined \def \showDOI       #1{#1}\fi
\ifx \showISBNx    \undefined \def \showISBNx     #1{\unskip}     \fi
\ifx \showISBNxiii \undefined \def \showISBNxiii  #1{\unskip}     \fi
\ifx \showISSN     \undefined \def \showISSN      #1{\unskip}     \fi
\ifx \showLCCN     \undefined \def \showLCCN      #1{\unskip}     \fi
\ifx \shownote     \undefined \def \shownote      #1{#1}          \fi
\ifx \showarticletitle \undefined \def \showarticletitle #1{#1}   \fi
\ifx \showURL      \undefined \def \showURL       {\relax}        \fi
\providecommand\bibfield[2]{#2}
\providecommand\bibinfo[2]{#2}
\providecommand\natexlab[1]{#1}
\providecommand\showeprint[2][]{arXiv:#2}

\bibitem[\protect\citeauthoryear{Abadi, Agarwal, Barham, Brevdo, Chen, Citro,
  Corrado, Davis, Dean, Devin, Ghemawat, Goodfellow, Harp, Irving, Isard, Jia,
  Jozefowicz, Kaiser, Kudlur, Levenberg, Man\'{e}, Monga, Moore, Murray, Olah,
  Schuster, Shlens, Steiner, Sutskever, Talwar, Tucker, Vanhoucke, Vasudevan,
  Vi\'{e}gas, Vinyals, Warden, Wattenberg, Wicke, Yu, and Zheng}{Abadi
  et~al\mbox{.}}{2015}]%
        {Abad15a}
\bibfield{author}{\bibinfo{person}{Mart\'{\i}n Abadi}, \bibinfo{person}{Ashish
  Agarwal}, \bibinfo{person}{Paul Barham}, \bibinfo{person}{Eugene Brevdo},
  \bibinfo{person}{Zhifeng Chen}, \bibinfo{person}{Craig Citro},
  \bibinfo{person}{Greg~S. Corrado}, \bibinfo{person}{Andy Davis},
  \bibinfo{person}{Jeffrey Dean}, \bibinfo{person}{Matthieu Devin},
  \bibinfo{person}{Sanjay Ghemawat}, \bibinfo{person}{Ian Goodfellow},
  \bibinfo{person}{Andrew Harp}, \bibinfo{person}{Geoffrey Irving},
  \bibinfo{person}{Michael Isard}, \bibinfo{person}{Yangqing Jia},
  \bibinfo{person}{Rafal Jozefowicz}, \bibinfo{person}{Lukasz Kaiser},
  \bibinfo{person}{Manjunath Kudlur}, \bibinfo{person}{Josh Levenberg},
  \bibinfo{person}{Dandelion Man\'{e}}, \bibinfo{person}{Rajat Monga},
  \bibinfo{person}{Sherry Moore}, \bibinfo{person}{Derek Murray},
  \bibinfo{person}{Chris Olah}, \bibinfo{person}{Mike Schuster},
  \bibinfo{person}{Jonathon Shlens}, \bibinfo{person}{Benoit Steiner},
  \bibinfo{person}{Ilya Sutskever}, \bibinfo{person}{Kunal Talwar},
  \bibinfo{person}{Paul Tucker}, \bibinfo{person}{Vincent Vanhoucke},
  \bibinfo{person}{Vijay Vasudevan}, \bibinfo{person}{Fernanda Vi\'{e}gas},
  \bibinfo{person}{Oriol Vinyals}, \bibinfo{person}{Pete Warden},
  \bibinfo{person}{Martin Wattenberg}, \bibinfo{person}{Martin Wicke},
  \bibinfo{person}{Yuan Yu}, {and} \bibinfo{person}{Xiaoqiang Zheng}.}
  \bibinfo{year}{2015}\natexlab{}.
\newblock \bibinfo{title}{{TensorFlow}: Large-Scale Machine Learning on
  Heterogeneous Systems}.
\newblock
\newblock
\urldef\tempurl%
\url{https://www.tensorflow.org/}
\showURL{%
\tempurl}
\newblock
\shownote{Software available from tensorflow.org.}


\bibitem[\protect\citeauthoryear{Amiriparian, Freitag, Cummins, and
  Schuller}{Amiriparian et~al\mbox{.}}{2017}]%
        {Amir17a}
\bibfield{author}{\bibinfo{person}{Shahin Amiriparian},
  \bibinfo{person}{Michael Freitag}, \bibinfo{person}{Nicholas Cummins}, {and}
  \bibinfo{person}{Bj{\"o}rn Schuller}.} \bibinfo{year}{2017}\natexlab{}.
\newblock \showarticletitle{Sequence to sequence autoencoders for unsupervised
  representation learning from audio}. In \bibinfo{booktitle}{\emph{Proceedings
  of the Detection and Classification of Acoustic Scenes and Events 2017
  Workshop}}. \bibinfo{pages}{17--21}.
\newblock


\bibitem[\protect\citeauthoryear{Atrey, Maddage, and Kankanhalli}{Atrey
  et~al\mbox{.}}{2006}]%
        {Atrey2006}
\bibfield{author}{\bibinfo{person}{Pradeep~K Atrey}, \bibinfo{person}{Namunu~C
  Maddage}, {and} \bibinfo{person}{Mohan~S Kankanhalli}.}
  \bibinfo{year}{2006}\natexlab{}.
\newblock \showarticletitle{Audio based event detection for multimedia
  surveillance}. In \bibinfo{booktitle}{\emph{2006 IEEE International
  Conference on Acoustics Speech and Signal Processing Proceedings}},
  Vol.~\bibinfo{volume}{5}. IEEE, \bibinfo{pages}{V--V}.
\newblock


\bibitem[\protect\citeauthoryear{Bayram, Duman, and Ince}{Bayram
  et~al\mbox{.}}{2021}]%
        {bayram2021real}
\bibfield{author}{\bibinfo{person}{Bar{\i}{\c{s}} Bayram},
  \bibinfo{person}{Taha~Berkay Duman}, {and} \bibinfo{person}{G{\"o}khan
  Ince}.} \bibinfo{year}{2021}\natexlab{}.
\newblock \showarticletitle{Real time detection of acoustic anomalies in
  industrial processes using sequential autoencoders}.
\newblock \bibinfo{journal}{\emph{Expert Systems}} \bibinfo{volume}{38},
  \bibinfo{number}{1} (\bibinfo{year}{2021}), \bibinfo{pages}{e12564}.
\newblock


\bibitem[\protect\citeauthoryear{Chachada and Kuo}{Chachada and Kuo}{2014}]%
        {Chachada2014}
\bibfield{author}{\bibinfo{person}{Sachin Chachada} {and}
  \bibinfo{person}{C.-C.~Jay Kuo}.} \bibinfo{year}{2014}\natexlab{}.
\newblock \showarticletitle{Environmental sound recognition: a survey}.
\newblock \bibinfo{journal}{\emph{{APSIPA} Transactions on Signal and
  Information Processing}}  \bibinfo{volume}{3} (\bibinfo{year}{2014}).
\newblock
\urldef\tempurl%
\url{https://doi.org/10.1017/atsip.2014.12}
\showDOI{\tempurl}


\bibitem[\protect\citeauthoryear{Chollet et~al\mbox{.}}{Chollet
  et~al\mbox{.}}{2015}]%
        {Chol15a}
\bibfield{author}{\bibinfo{person}{Fran\c{c}ois Chollet} {et~al\mbox{.}}}
  \bibinfo{year}{2015}\natexlab{}.
\newblock \bibinfo{title}{Keras}.
\newblock
\newblock
\urldef\tempurl%
\url{https://keras.io}
\showURL{%
\tempurl}


\bibitem[\protect\citeauthoryear{Copeland}{Copeland}{2010}]%
        {CRS10b}
\bibfield{author}{\bibinfo{person}{Claudia Copeland}.}
  \bibinfo{year}{2010}\natexlab{}.
\newblock \bibinfo{booktitle}{\emph{Terrorism and Security Issues Facing the
  Water Infrastructure Sector}}.
\newblock \bibinfo{publisher}{Congressional Research Service}.
\newblock


\bibitem[\protect\citeauthoryear{Duan, Yang, Ma, Yang, and Li}{Duan
  et~al\mbox{.}}{2014}]%
        {Duan14a}
\bibfield{author}{\bibinfo{person}{Rui Duan}, \bibinfo{person}{Kunde Yang},
  \bibinfo{person}{Yuanliang Ma}, \bibinfo{person}{Qiulong Yang}, {and}
  \bibinfo{person}{Hui Li}.} \bibinfo{year}{2014}\natexlab{}.
\newblock \showarticletitle{Moving source localization with a single hydrophone
  using multipath time delays in the deep ocean}.
\newblock \bibinfo{journal}{\emph{The Journal of the Acoustical Society of
  America}} \bibinfo{volume}{136}, \bibinfo{number}{2} (\bibinfo{date}{August}
  \bibinfo{year}{2014}), \bibinfo{pages}{159--165}.
\newblock
\urldef\tempurl%
\url{https://doi.org/10.1121/1.4890664}
\showDOI{\tempurl}


\bibitem[\protect\citeauthoryear{Fischbach, Strohschein, Bunte, Stork,
  Faeskorn-Woyke, Moriz, and Bartz-Beielstein}{Fischbach et~al\mbox{.}}{2020}]%
        {Fisc20a}
\bibfield{author}{\bibinfo{person}{Andreas Fischbach}, \bibinfo{person}{Jan
  Strohschein}, \bibinfo{person}{Andreas Bunte}, \bibinfo{person}{Jörg Stork},
  \bibinfo{person}{Heide Faeskorn-Woyke}, \bibinfo{person}{Natalia Moriz},
  {and} \bibinfo{person}{Thomas Bartz-Beielstein}.}
  \bibinfo{year}{2020}\natexlab{}.
\newblock \showarticletitle{CAAI - A cognitive architecture to introduce
  artificial intelligence in cyber-physical production systems}.
\newblock \bibinfo{journal}{\emph{The International Journal of Advanced
  Manufacturing Technology}} \bibinfo{volume}{111}, \bibinfo{number}{1-2}
  (\bibinfo{date}{oct} \bibinfo{year}{2020}), \bibinfo{pages}{609--626}.
\newblock
\urldef\tempurl%
\url{https://doi.org/10.1007/s00170-020-06094-z}
\showDOI{\tempurl}


\bibitem[\protect\citeauthoryear{Freitag, Amiriparian, Pugachevskiy, Cummins,
  and Schuller}{Freitag et~al\mbox{.}}{2018}]%
        {Frei18a}
\bibfield{author}{\bibinfo{person}{Michael Freitag}, \bibinfo{person}{Shahin
  Amiriparian}, \bibinfo{person}{Sergey Pugachevskiy},
  \bibinfo{person}{Nicholas Cummins}, {and} \bibinfo{person}{Bj\"{o}rn
  Schuller}.} \bibinfo{year}{2018}\natexlab{}.
\newblock \showarticletitle{auDeep: Unsupervised Learning of Representations
  from Audio with Deep Recurrent Neural Networks}.
\newblock \bibinfo{journal}{\emph{Journal of Machine Learning Research}}
  \bibinfo{volume}{18}, \bibinfo{number}{173} (\bibinfo{year}{2018}),
  \bibinfo{pages}{1--5}.
\newblock
\urldef\tempurl%
\url{http://jmlr.org/papers/v18/17-406.html}
\showURL{%
\tempurl}


\bibitem[\protect\citeauthoryear{Kingma and Ba}{Kingma and Ba}{2015}]%
        {King15a}
\bibfield{author}{\bibinfo{person}{Diederik~P. Kingma} {and}
  \bibinfo{person}{Jimmy Ba}.} \bibinfo{year}{2015}\natexlab{}.
\newblock \showarticletitle{Adam: {A} Method for Stochastic Optimization}. In
  \bibinfo{booktitle}{\emph{3rd International Conference on Learning
  Representations, {ICLR} 2015, San Diego, CA, USA, May 7-9, 2015, Conference
  Track Proceedings}}, \bibfield{editor}{\bibinfo{person}{Yoshua Bengio} {and}
  \bibinfo{person}{Yann LeCun}} (Eds.).
\newblock
\urldef\tempurl%
\url{http://arxiv.org/abs/1412.6980}
\showURL{%
\tempurl}


\bibitem[\protect\citeauthoryear{Li, Yang, and Hu}{Li et~al\mbox{.}}{2019}]%
        {li19b}
\bibfield{author}{\bibinfo{person}{Bin Li}, \bibinfo{person}{Jie Yang}, {and}
  \bibinfo{person}{Dexiu Hu}.} \bibinfo{year}{2019}\natexlab{}.
\newblock \showarticletitle{Dam monitoring data analysis methods: A literature
  review}.
\newblock \bibinfo{journal}{\emph{Structural Control and Health Monitoring}}
  \bibinfo{volume}{27}, \bibinfo{number}{3} (\bibinfo{year}{2019}).
\newblock
\urldef\tempurl%
\url{https://doi.org/10.1002/stc.2501}
\showDOI{\tempurl}


\bibitem[\protect\citeauthoryear{Marchi, Vesperini, Eyben, Squartini, and
  Schuller}{Marchi et~al\mbox{.}}{2015}]%
        {marchi2015novel}
\bibfield{author}{\bibinfo{person}{Erik Marchi}, \bibinfo{person}{Fabio
  Vesperini}, \bibinfo{person}{Florian Eyben}, \bibinfo{person}{Stefano
  Squartini}, {and} \bibinfo{person}{Bj{\"o}rn Schuller}.}
  \bibinfo{year}{2015}\natexlab{}.
\newblock \showarticletitle{A novel approach for automatic acoustic novelty
  detection using a denoising autoencoder with bidirectional LSTM neural
  networks}. In \bibinfo{booktitle}{\emph{Proceedings 40th IEEE International
  Conference on Acoustics, Speech, and Signal Processing, ICASSP 2015}}.
  \bibinfo{pages}{5--pages}.
\newblock


\bibitem[\protect\citeauthoryear{Multiconsult}{Multiconsult}{2021}]%
        {multiconsult21b}
\bibfield{author}{\bibinfo{person}{Multiconsult}.}
  \bibinfo{year}{2021}\natexlab{}.
\newblock \bibinfo{booktitle}{\emph{Review of dam monitoring and data
  management techniques}}.
\newblock \bibinfo{publisher}{Multiconsult}.
\newblock


\bibitem[\protect\citeauthoryear{Normand}{Normand}{2019}]%
        {CRS19b}
\bibfield{author}{\bibinfo{person}{Anna Normand}.}
  \bibinfo{year}{2019}\natexlab{}.
\newblock \bibinfo{booktitle}{\emph{Dam Safety Overview and the Federal Role}}.
\newblock \bibinfo{publisher}{Congressional Research Service}.
\newblock


\bibitem[\protect\citeauthoryear{Radhakrishnan, Divakaran, and
  Smaragdis}{Radhakrishnan et~al\mbox{.}}{2005}]%
        {Radhakrishnan2005}
\bibfield{author}{\bibinfo{person}{Regunathan Radhakrishnan},
  \bibinfo{person}{Ajay Divakaran}, {and} \bibinfo{person}{A Smaragdis}.}
  \bibinfo{year}{2005}\natexlab{}.
\newblock \showarticletitle{Audio analysis for surveillance applications}. In
  \bibinfo{booktitle}{\emph{IEEE Workshop on Applications of Signal Processing
  to Audio and Acoustics, 2005.}} IEEE, \bibinfo{pages}{158--161}.
\newblock


\bibitem[\protect\citeauthoryear{Shipps and Abraham}{Shipps and
  Abraham}{2004}]%
        {Shipps2004}
\bibfield{author}{\bibinfo{person}{J.C. Shipps} {and} \bibinfo{person}{B.M.
  Abraham}.} \bibinfo{year}{2004}\natexlab{}.
\newblock \showarticletitle{The use of vector sensors for underwater port and
  waterway security}. In \bibinfo{booktitle}{\emph{{ISA}/{IEEE} Sensors for
  Industry Conference, 2004. Proceedings the}}. \bibinfo{publisher}{{IEEE}},
  \bibinfo{address}{New Orleans, LA, USA}.
\newblock
\urldef\tempurl%
\url{https://doi.org/10.1109/sficon.2004.1287125}
\showDOI{\tempurl}


\bibitem[\protect\citeauthoryear{Skarsoulis, Piperakis, Kalogerakis, Orfanakis,
  Papadakis, Dosso, and Frantzis}{Skarsoulis et~al\mbox{.}}{2018}]%
        {Skar18a}
\bibfield{author}{\bibinfo{person}{Emmanuel Skarsoulis},
  \bibinfo{person}{George Piperakis}, \bibinfo{person}{Michael Kalogerakis},
  \bibinfo{person}{Emmanuel Orfanakis}, \bibinfo{person}{Panagiotis Papadakis},
  \bibinfo{person}{Stan. Dosso}, {and} \bibinfo{person}{Alexandros Frantzis}.}
  \bibinfo{year}{2018}\natexlab{}.
\newblock \showarticletitle{Underwater Acoustic Pulsed Source Localization with
  a Pair of Hydrophones}.
\newblock \bibinfo{journal}{\emph{Remote Sensing}} \bibinfo{volume}{10},
  \bibinfo{number}{6} (\bibinfo{date}{jun} \bibinfo{year}{2018}),
  \bibinfo{pages}{883}.
\newblock
\urldef\tempurl%
\url{https://doi.org/10.3390/rs10060883}
\showDOI{\tempurl}


\bibitem[\protect\citeauthoryear{Sutskever, Vinyals, and Le}{Sutskever
  et~al\mbox{.}}{2014}]%
        {Suts14a}
\bibfield{author}{\bibinfo{person}{Ilya Sutskever}, \bibinfo{person}{Oriol
  Vinyals}, {and} \bibinfo{person}{Quoc~V. Le}.}
  \bibinfo{year}{2014}\natexlab{}.
\newblock \showarticletitle{Sequence to Sequence Learning with Neural
  Networks}. In \bibinfo{booktitle}{\emph{Proceedings of the 27th International
  Conference on Neural Information Processing Systems - Volume 2}} (Montreal,
  Canada) \emph{(\bibinfo{series}{NIPS'14})}. \bibinfo{publisher}{MIT Press},
  \bibinfo{address}{Cambridge, MA, USA}, \bibinfo{pages}{3104–3112}.
\newblock


\bibitem[\protect\citeauthoryear{Thode}{Thode}{2004}]%
        {Thod04a}
\bibfield{author}{\bibinfo{person}{Aaron Thode}.}
  \bibinfo{year}{2004}\natexlab{}.
\newblock \showarticletitle{Tracking sperm whale (Physeter macrocephalus) dive
  profiles using a towed passive acoustic array}.
\newblock \bibinfo{journal}{\emph{The Journal of the Acoustical Society of
  America}} \bibinfo{volume}{116}, \bibinfo{number}{1} (\bibinfo{date}{jul}
  \bibinfo{year}{2004}), \bibinfo{pages}{245--253}.
\newblock
\urldef\tempurl%
\url{https://doi.org/10.1121/1.1758972}
\showDOI{\tempurl}


\bibitem[\protect\citeauthoryear{Wienand}{Wienand}{2019}]%
        {BBK19b}
\bibfield{author}{\bibinfo{person}{Ina Wienand}.}
  \bibinfo{year}{2019}\natexlab{}.
\newblock \bibinfo{booktitle}{\emph{Sicherheit der Trinkwasserversorgung}}.
\newblock \bibinfo{publisher}{Bundesamt f{\"u}r Bev{\"o}elkerungsschutz und
  Katastrophenhilfe}.
\newblock


\end{thebibliography}

%
%
%
%
%
%
%
%

\end{document}